\documentclass[letterpaper]{article}
\usepackage[preprint]{aaai2027}
\usepackage[hyphens]{url}
\usepackage{graphicx}
\usepackage{natbib}
\usepackage{caption}
\usepackage{amsmath,amssymb}
\usepackage{booktabs,array}
\usepackage{multirow}
\usepackage{calc}
\title{Pipette: An Embodied Simulation Platform, Benchmark, and Success-Verified Simulation Augmentation Framework for Wet-Lab Robotics}
\author{
Zhe~Liu\equalcontrib\textsuperscript{\rm 1,\rm 2},
Huanbo~Jin\equalcontrib\textsuperscript{\rm 1,\rm 2},
Zhaohui~Du\equalcontrib\textsuperscript{\rm 1,\rm 2},
Zhe~Wang\corresponding\textsuperscript{\rm 1,\rm 2},\\
Dongzhan~Zhou\textsuperscript{\rm 4},
Minting~Pan\textsuperscript{\rm 4},
He~Xu\textsuperscript{\rm 2},
Peijia~Li\textsuperscript{\rm 2},\\
Jiaming~Gu\textsuperscript{\rm 1,\rm 2},
Quan~Lu\textsuperscript{\rm 1,\rm 2},
Qi~Wang\textsuperscript{\rm 3},
Bin~Ji\textsuperscript{\rm 1,\rm 2},
Ting~Xiao\textsuperscript{\rm 1,\rm 2}
}
\affiliations{
\textsuperscript{\rm 1}Key Laboratory of Smart Manufacturing in Energy Chemical Process, Ministry of Education,\\
East China University of Science and Technology, Shanghai, China\\
\textsuperscript{\rm 2}Department of Computer Science and Engineering,\\
East China University of Science and Technology, Shanghai, China\\
\textsuperscript{\rm 3}Department of Laboratory Medicine, Ruijin Hospital,\\
Shanghai Jiao Tong University School of Medicine, Shanghai, China\\
\textsuperscript{\rm 4}AI for Science Center, Shanghai AI Laboratory, Shanghai, CN\\
wangzhe@ecust.edu.cn
}

\begin{document}
\maketitle

\begin{abstract}
Biomedical laboratory robots can improve experimental reproducibility, throughput, and safety, but their development is constrained by costly demonstrations and the lack of simulation environments tailored to biomedical instruments and consumables. We present Pipette, an embodied simulation platform, benchmark, and simulation-augmentation framework for biomedical laboratory robotics. Its core contribution is a success-verified simulation augmentation pipeline that replays limited human demonstrations under controlled variations in lighting, camera pose, execution speed, and robot actions. Each replay regenerates synchronized observations, robot states, and actions through physical simulation, after which Pipette retains only episodes that satisfy task-specific success criteria. This process preserves physical and temporal consistency while filtering unsuccessful or dynamically invalid augmented trajectories. The platform further provides more than 100 open-source and re-editable biomedical laboratory assets, supports three robotic-arm embodiments, and uses a unified interface for task construction, data collection, training, and evaluation. We establish a 12-task benchmark covering consumable handling, cultureware manipulation, instrument operation, and precision placement. Using only 30 manual demonstrations per task, augmentation improves 26 of 36 policy--task combinations, raising the average success rate of SmolVLA from 40.4\% to 71.8\% and \(\pi\)0 from 37.3\% to 44.1\%. ACT also improves on most tasks, although gains remain sensitive to contact-intensive precision placement. Pipette further supports language-assisted task registration, providing a specialized and reproducible testbed for robot learning from limited demonstrations in biomedical laboratories.
\end{abstract}

\begin{center}
\textbf{Project page:} \url{https://github.com/hbhuiyou/Pipette}
\end{center}

\section{Introduction}\label{introduction}

Biomedical laboratory automation is important for improving the reproducibility and safety of life science experiments. Automated systems are valuable when operations must be repeated consistently across many samples~\citep{holland2020automation}. However, biomedical protocols often impose strict timing requirements, and instruments from different manufacturers can be difficult to coordinate~\citep{rupp2024benefit}. Self-driving laboratories seek to connect experimental design with automated execution and iterative optimization~\citep{tobias2025autonomous}, while agentic AI extends this vision by interpreting experimental goals and coordinating procedures~\citep{hartung2025agentic}. Most existing systems nevertheless remain coupled to fixed instruments and predefined scripts. They are less suited to operations requiring visual perception and adaptive interaction, such as grasping a small consumable or moving an instrument component. Biomedical laboratory automation therefore needs embodied environments that connect experimental instructions with robot perception and action.

Recent robot foundation models and vision-language-action (VLA) models support flexible manipulation. RT-1 and RT-2 demonstrate that policies trained on large robotic datasets can follow language instructions across multiple tasks~\citep{brohan2022rt1,brohan2023rt2}. Open X-Embodiment shows the value of standardized data collected from different robots~\citep{openx2024}. BioProVLA-Agent explores protocol interpretation, visual verification, and VLA-based execution in biomedical laboratories~\citep{du2026bioprovla}. These advances still depend on demonstrations. Collecting trajectories requires physical hardware, scene preparation, and human supervision~\citep{khazatsky2024droid}. The burden is greater in biomedical laboratories, where errors may damage samples, contaminate the workspace, or interrupt an experiment.

Simulation can reduce this burden and support reproducible policy evaluation. Existing platforms support general manipulation, lifelong learning, demonstration synthesis, and household robotics~\citep{james2020rlbench,liu2023libero,mandlekar2023mimicgen,nasiriany2024robocasa}, but their assets and task abstractions are rarely designed for biomedical laboratories. A suitable platform must represent consumables alongside instruments whose movable parts participate in a task. It must also capture the narrow tolerances involved in grasping a tube or placing an object within an instrument. Data augmentation creates another challenge. Image transformations alter appearance without updating robot states or actions. Physical trajectory perturbations provide more meaningful variation, but the modified motion may no longer complete the intended operation. Without automatic verification, failed replays can enter the training set as incorrect demonstrations.

Here, we introduce Pipette, an embodied simulation platform, benchmark, and simulation-augmentation framework for biomedical laboratory robotics. Pipette provides reusable digital assets for consumables, equipment, and interactive instruments. A common task representation links scene configuration to robot initialization, camera settings, language instructions, and success criteria, allowing one definition to support data collection, training, and evaluation. Its central component is a success-verified simulation augmentation pipeline. Pipette replays limited human demonstrations under controlled environmental and action perturbations, regenerates synchronized observations, states, and actions, and retains only episodes that satisfy task-specific evaluators. This expands training data while preserving consistency between robot behavior and task outcome.

We instantiate Pipette with more than 100 editable assets and 12 biomedical laboratory tasks. Through a shared interface, we train and evaluate ACT, SmolVLA, and \(\pi\)0 using 30 demonstrations per task. Simulation augmentation improves 26 of the 36 policy--task combinations, with particularly strong gains for SmolVLA. Its effectiveness also depends on policy architecture and task geometry, especially when execution requires stable contact and precise release.

Our contributions are summarized as follows:
\begin{itemize}
    \item We develop an extensible biomedical laboratory simulation platform containing more than 100 editable USD assets and supporting three robotic-arm embodiments.
    \item We introduce a language-assisted task registration mechanism that organizes scene, robot, sensor, instruction, and evaluator information through a shared representation.
    \item We propose success-verified simulation augmentation, which replays demonstrations under controlled perturbations and filters the resulting episodes through task-specific success checks.
    \item We establish a 12-task benchmark for evaluating ACT, SmolVLA, and \(\pi\)0 under consistent data, control, and evaluation settings.
\end{itemize}

\section{Related Work}\label{related-work}

\subsection{Laboratory Automation and Scientific Embodied AI}\label{laboratory-automation-and-scientific-embodied-ai}

Recent biological laboratory automation has evolved from isolated liquid-handling scripts toward integrated systems that combine protocol formalization, robotic execution, and data-driven experimental optimization. \citet{stephenson2023physical} reviewed physical laboratory automation in synthetic biology and identified the limited connectivity between design-build-test-learn modules as a major barrier to broader adoption. To improve protocol reuse and implementation, \citet{anhel2023lap} proposed the Laboratory Automation Protocol format, which standardizes script-based automation protocols and supports modular workflow composition. Beyond protocol sharing, \citet{jiang2024protocode} introduced ProtoCode to convert PCR procedures from scientific publications into machine-readable protocols, showing the potential of language models for reducing manual protocol curation. More recently, \citet{rapp2024selfdriving} demonstrated a self-driving platform for autonomous protein engineering, while \citet{fushimi2025autonomous} developed an autonomous laboratory system that integrates culturing, preprocessing, measurement, analysis, and hypothesis formulation for biotechnology experiments. These studies indicate that biological laboratory automation is moving toward more autonomous and closed-loop experimental workflows. However, existing systems are still largely constrained by fixed instruments, predefined scripts, and task-specific workflows, making it difficult to support flexible manipulation of diverse labware, visual state changes, and multi-step wet-lab operations. This limitation motivates the need for more general embodied execution methods that can bridge natural experimental protocols and physical robotic manipulation.

\subsection{Vision-Language-Action Models for Robotic Manipulation}\label{vision-language-action-models-for-robotic-manipulation}

Vision-language-action models have recently emerged as a promising paradigm for grounding visual observations and language instructions into executable robot actions. \citet{zhao2023act} proposed ACT, an action-chunking transformer that predicts short action sequences for fine-grained manipulation, providing a strong imitation-learning baseline although it is not a VLA foundation model. \citet{kim2024openvla} introduced OpenVLA, an open-source 7B vision-language-action model trained on large-scale robot demonstrations for visuomotor control. \citet{black2024pi0} proposed \(\pi\)0, which combines pretrained vision-language backbones with a flow-matching action expert to generate continuous robot actions across different embodiments. Black et al. \citeyearpar{black2025pi05} further developed \ensuremath{\pi}0.5 to improve open-world generalization through co-training with heterogeneous robotic, semantic, and web-scale supervision. \citet{shukor2025smolvla} introduced SmolVLA, which focuses on affordable and efficient robotics by reducing the training and deployment cost of VLA models while maintaining competitive manipulation performance. These studies demonstrate the potential of generalist policies for language-conditioned robotic control. However, current VLA models still face challenges in high-precision contact-rich manipulation, domain transfer, visual ambiguity, and data efficiency. These limitations are especially critical in biological wet-lab scenarios, where transparent labware, small targets, constrained workspaces, and expensive real-world data collection make direct training and evaluation difficult.

\begin{table*}[!t]
\centering
\begingroup
\small
\begin{tabular}{@{}
  >{\raggedright\arraybackslash}m{(\linewidth - 16\tabcolsep) * \real{0.125}}
  >{\raggedright\arraybackslash}m{(\linewidth - 16\tabcolsep) * \real{0.180}}
  >{\centering\arraybackslash}m{(\linewidth - 16\tabcolsep) * \real{0.055}}
  >{\centering\arraybackslash}m{(\linewidth - 16\tabcolsep) * \real{0.100}}
  >{\centering\arraybackslash}m{(\linewidth - 16\tabcolsep) * \real{0.090}}
  >{\centering\arraybackslash}m{(\linewidth - 16\tabcolsep) * \real{0.130}}
  >{\centering\arraybackslash}m{(\linewidth - 16\tabcolsep) * \real{0.110}}
  >{\centering\arraybackslash}m{(\linewidth - 16\tabcolsep) * \real{0.105}}
  >{\centering\arraybackslash}m{(\linewidth - 16\tabcolsep) * \real{0.105}}@{}}
\toprule\noalign{}
\begin{minipage}[c]{\linewidth}\raggedright
Benchmark
\end{minipage} & \begin{minipage}[c]{\linewidth}\raggedright
Target Domain
\end{minipage} & \begin{minipage}[c]{\linewidth}\centering
Tasks
\end{minipage} & \begin{minipage}[c]{\linewidth}\centering
Interactive Instruments
\end{minipage} & \begin{minipage}[c]{\linewidth}\centering
NL-driven
\end{minipage} & \begin{minipage}[c]{\linewidth}\centering
Ext. Wet-Lab Reg.
\end{minipage} & \begin{minipage}[c]{\linewidth}\centering
Data Augmentation
\end{minipage} & \begin{minipage}[c]{\linewidth}\centering
Data Synthesis
\end{minipage} & \begin{minipage}[c]{\linewidth}\centering
VLA Train \& Eval
\end{minipage} \\
\midrule\noalign{}
Meta-World & Domain-agnostic & 50 & \ensuremath{\times} & \ensuremath{\times} & \ensuremath{\times} & \ensuremath{\times} & \ensuremath{\checkmark} & \ensuremath{\times} \\
RoboSuite & Domain-agnostic & 9 & \ensuremath{\times} & \ensuremath{\times} & \ensuremath{\times} & \ensuremath{\times} & \ensuremath{\times} & \ensuremath{\times} \\
Factory & Industrial manipulation & 8 & \ensuremath{\times} & \ensuremath{\times} & \ensuremath{\times} & \ensuremath{\times} & \ensuremath{\times} & \ensuremath{\times} \\
ManiSkill2 & Domain-agnostic & 20 & \ensuremath{\times} & \ensuremath{\times} & \ensuremath{\times} & \ensuremath{\times} & \ensuremath{\checkmark} & \ensuremath{\times} \\
RoboTwin 2.0 & Domain-agnostic & 50 & \ensuremath{\times} & \ensuremath{\times} & \ensuremath{\times} & \ensuremath{\checkmark} & \ensuremath{\checkmark} & \ensuremath{\checkmark} \\
Chemistry3D & Chemistry & 5 & \ensuremath{\times} & \ensuremath{\times} & Limited & \ensuremath{\times} & \ensuremath{\checkmark} & \ensuremath{\times} \\
AutoBio & Biomedical wet lab & 16 & \ensuremath{\checkmark} & \ensuremath{\times} & Limited & \ensuremath{\times} & \ensuremath{\checkmark} & \ensuremath{\checkmark} \\
LabUtopia & Scientific lab & 30 & \ensuremath{\checkmark} & Limited & Limited & \ensuremath{\times} & \ensuremath{\checkmark} & \ensuremath{\checkmark} \\
Pipette(Ours) & Biomedical wet lab & 12 & \ensuremath{\checkmark} & \ensuremath{\checkmark} & \ensuremath{\checkmark} & \ensuremath{\checkmark} & \ensuremath{\checkmark} & \ensuremath{\checkmark} \\
\bottomrule
\end{tabular}
\endgroup
\caption{Comparison of existing robotic simulation platforms in terms of domain specificity, interactive instrumentation, language-driven task construction, extensible laboratory assets, data augmentation, data synthesis, VLA training and evaluation}
\label{tab:platform-comparison}
\end{table*}

\subsection{Embodied AI and Robotic Simulation Platforms}\label{embodied-ai-and-robotic-simulation-platforms}

Robotic simulation platforms are essential for scalable data generation, reproducible policy evaluation, and safe development of embodied AI systems before real-world deployment. \citet{yu2019metaworld} introduced Meta-World, a 50-task manipulation benchmark for multi-task and meta-reinforcement learning. \citet{zhu2020robosuite} developed robosuite, a modular MuJoCo-based framework for constructing reproducible robot-learning environments. \citet{narang2022factory} proposed Factory, a set of simulation methods and benchmark environments for contact-rich industrial assembly tasks, emphasizing precise insertion, fastening, and force-sensitive manipulation. \citet{gu2023maniskill2} proposed ManiSkill2, which expands simulation-based manipulation benchmarks by covering diverse object models, task families, robot embodiments, and observation modalities. Recent platforms have further moved beyond generic tabletop manipulation toward more specialized scenarios. \citet{li2024chemistry3d} introduced Chemistry3D, a chemistry-oriented robotic interaction environment that supports liquid flow, transparent objects, and reaction-state visualization. \citet{li2025labutopia} introduced LabUtopia, a broader scientific-laboratory simulation and hierarchical benchmark that extends beyond chemistry-specific interactions to cover more diverse lab scenes, instruments, and embodied-agent tasks. \citet{chen2025robotwin2} proposed RoboTwin 2.0, a scalable data-generation and benchmarking framework for robust bimanual manipulation. \citet{lan2025autobio} developed AutoBio to extend robotic simulation toward biological laboratory automation by modeling biology-grounded tasks, laboratory instruments, transparent materials, and VLA-oriented evaluation. However, as summarized in Table 1, existing simulation environments are still dominated by general-purpose manipulation, industrial assembly, or chemistry-oriented experiments, while biological wet-lab simulation remains less explored. Moreover, few platforms simultaneously support interactive instruments, natural-language or agent-driven task specification, extensible wet-lab task registration, data augmentation, data synthesis, and VLA-oriented training and evaluation. To address this gap, we propose Pipette, a biomedical wet-lab simulation platform designed for task construction, data generation, and evaluation of VLA-based biological robotic manipulation.

\section{Pipette Platform}\label{pipette-platform}

\subsection{Overview}\label{pipette-platform3.1-overview}

\begin{figure*}[!t]
\centering
\includegraphics[width=0.82\textwidth]{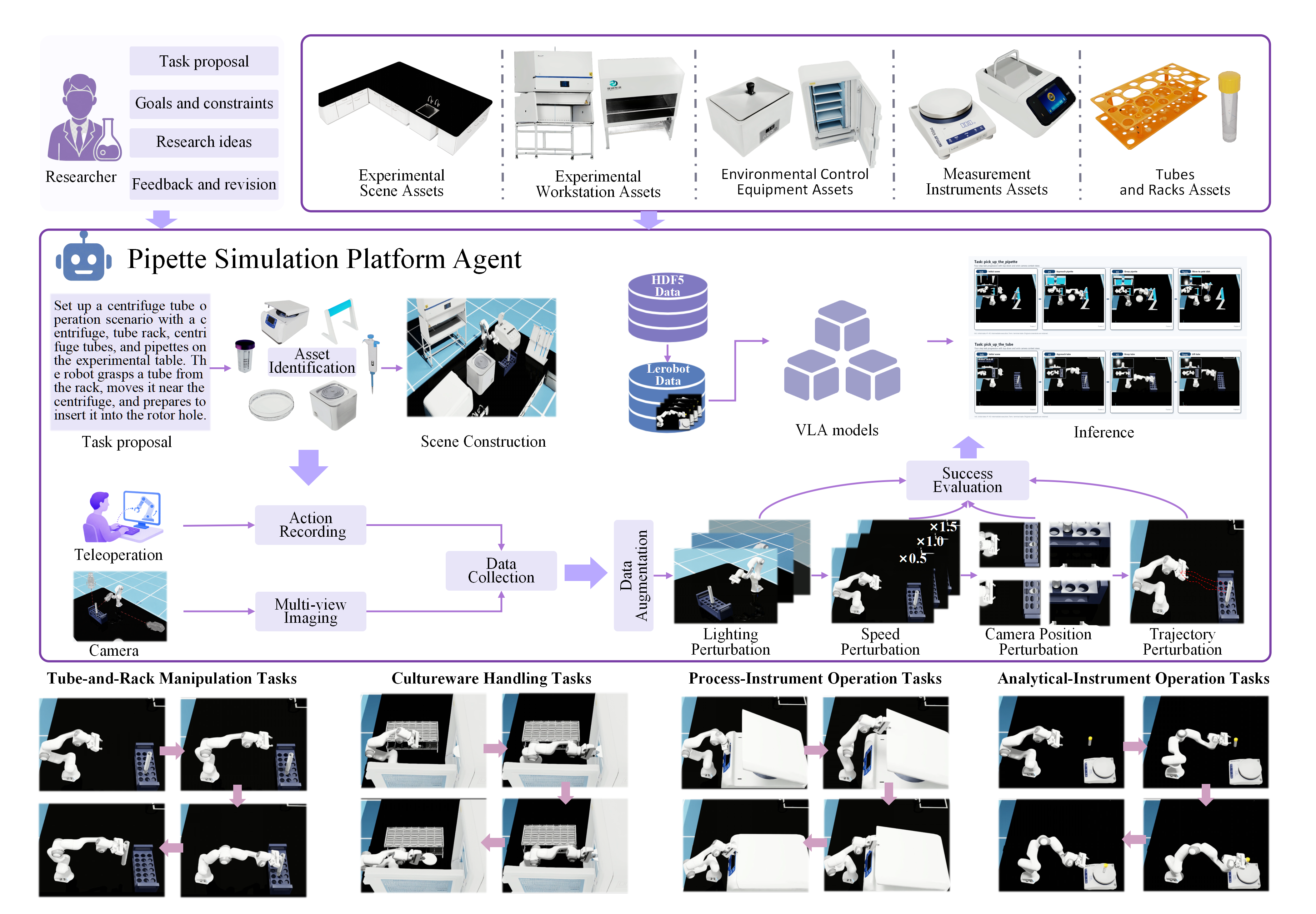}
\caption{Overview of the Pipette platform for wet-lab robotic simulation, augmentation, and evaluation}
\label{fig:platform-overview}
\end{figure*}

Existing wet-lab simulators provide limited interactive assets, inconsistent task and evaluation definitions, and few physically consistent mechanisms for expanding demonstrations. Pipette addresses these limitations with an extensible simulation, benchmark, and data-augmentation framework. It represents instruments and consumables as interactive USD assets with geometry, physical properties, collision structures, and task semantics, while a language-conditioned registration mechanism encodes scenes, robot and camera configurations, instructions, targets, and success criteria in a unified task format. This representation allows tasks to share data collection, augmentation, training, and evaluation interfaces.

Figure~\ref{fig:platform-overview} summarizes the pipeline, which integrates the asset library, task registration, demonstration collection, simulation augmentation, and VLA training and evaluation. Teleoperated or existing trajectories are replayed with lighting, camera, speed, and action perturbations to generate synchronized multi-view images, proprioceptive states, actions, and success labels; verified episodes are then converted to LeRobot format. Pipette provides 12 tasks spanning sample handling, instrument interaction, placement, lid operation, and target alignment. The benchmark uses three RGB views and robot proprioception as observations. The Franka Panda serves as the default robot, while an AgileX robotic arm is used to place an Erlenmeyer flask on a shaker. The unified setup supports controlled policy comparisons and failure-mode analysis.

\subsection{Problem Formulation}\label{problem-formulation}

We formulate each task \(\tau \in \mathcal{T}\) as language-conditioned closed-loop control with a fixed instruction \(\ell_{\tau}\). At time \(t\), the policy observes three RGB views and the robot proprioceptive state:

\begin{equation}
o_{t}=\left\{I_{t}^{top},I_{t}^{main},I_{t}^{wrist},s_{t}\right\},\qquad s_{t}\in\mathbb{R}^{8},
\end{equation}

where \(s_t\) contains seven joint positions and one gripper state. The policy outputs

\begin{equation}
a_{t}=\pi\left(o_{t},\ell_{\tau}\right)\in\mathbb{R}^{8},
\end{equation}

with seven joint targets and one gripper command. For each task, \(N_{\tau}\) evaluation rollouts produce binary outcomes \(S_{\tau}(\xi_{\tau}^{(i)})\in\{0,1\}\). The evaluator is used for comparison rather than direct optimization. The benchmark reports macro-average task success:

\begin{equation}
\hat{S}_{\tau}=\frac{1}{N_{\tau}}\sum_{i=1}^{N_{\tau}}S_{\tau}(\xi_{\tau}^{(i)}),\qquad \bar{S}=\frac{1}{|\mathcal{T}|}\sum_{\tau\in\mathcal{T}}\hat{S}_{\tau}.
\end{equation}

\subsection{Expandable Interactive Wet-Lab Assets}\label{expandable-interactive-wet-lab-assets}

\begin{figure}[!t]
\centering
\includegraphics[width=0.92\columnwidth]{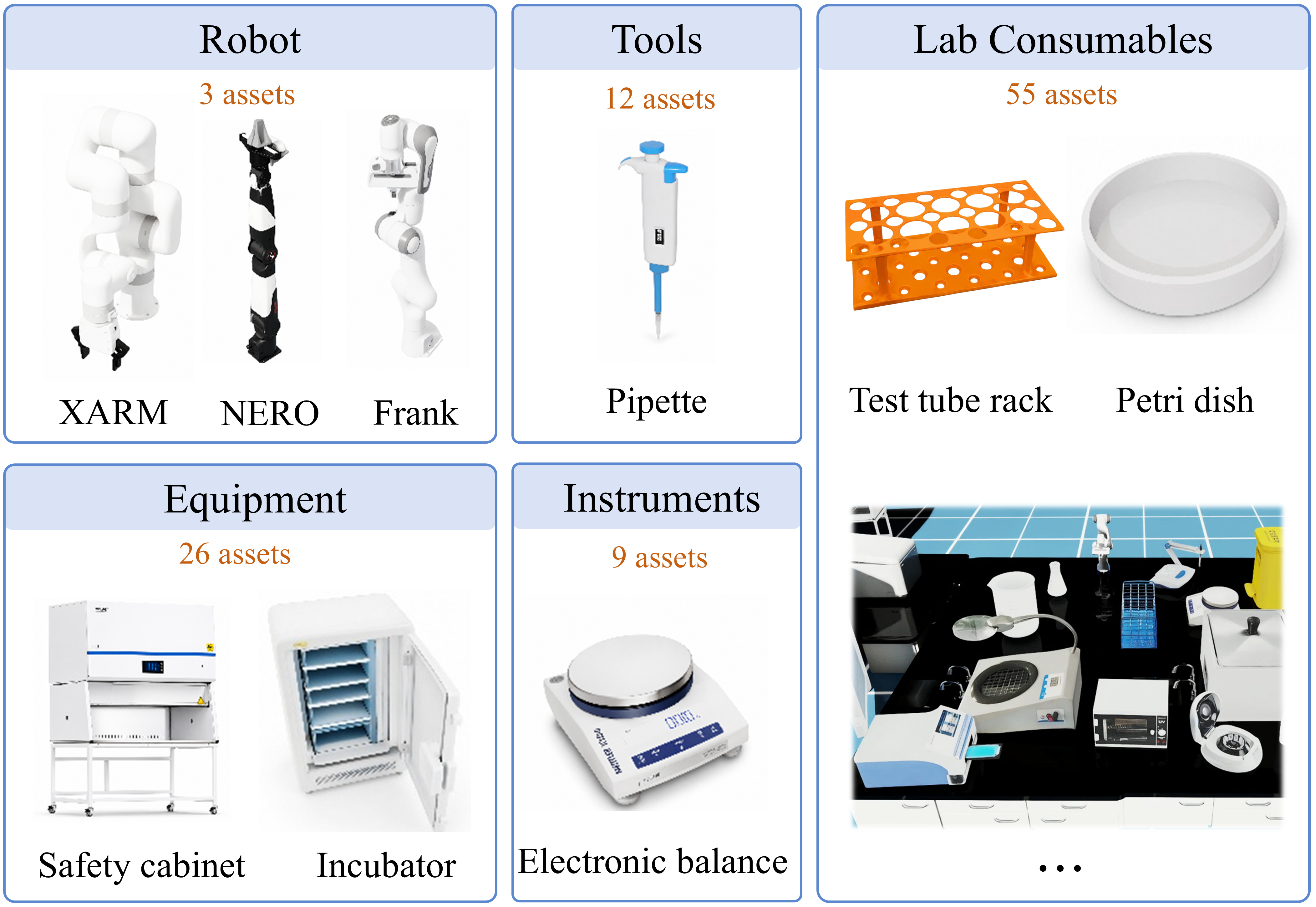}
\caption{Overview of the Pipette asset library. The platform includes robot embodiments, tools, equipment, lab consumables, and instruments for constructing interactive wet-lab simulation tasks.}
\label{fig:asset-library-overview}
\end{figure}

Wet-lab simulation requires assets that are both visually realistic and physically actionable. Pipette provides a unified library of reusable USD assets, including robot embodiments, tools, equipment, laboratory consumables, and instruments. As shown in Figure~\ref{fig:asset-library-overview}, the platform currently includes 3 robot embodiments, 12 tools, 26 equipment assets, 55 lab consumables, and 9 instruments, while the benchmark experiments use Franka Panda as the default robot unless otherwise specified.

Pipette represents instruments and consumables as interactive USD assets containing geometry, materials, collision bodies, physical properties, and task semantics. This common representation supports grasping, placement, device actuation, and task-level success evaluation.

\begin{figure*}[!t]
\centering
\includegraphics[width=0.82\textwidth]{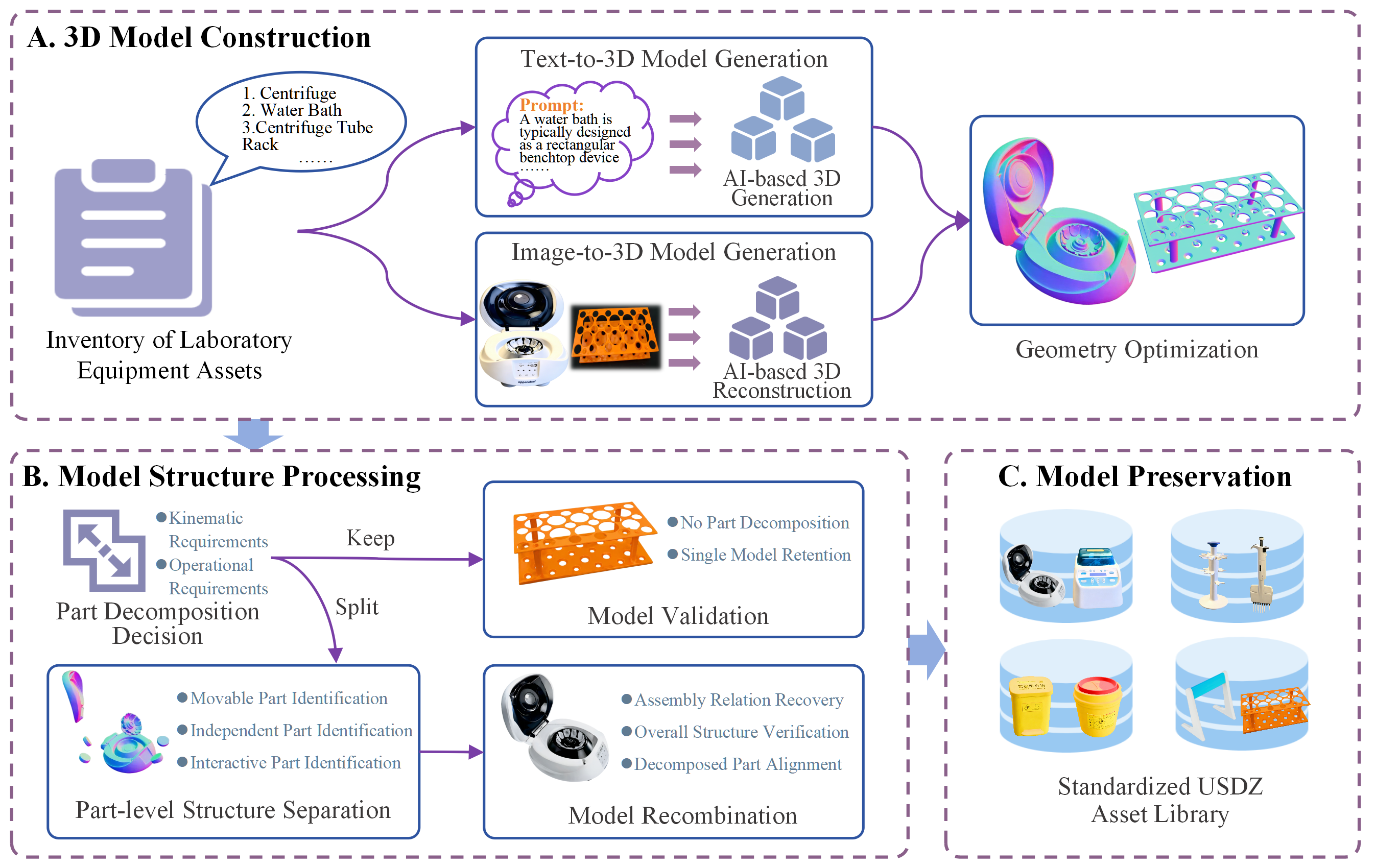}
\caption{Asset construction, structure processing, and preservation pipeline for interactive wet-lab digital assets}
\label{fig:asset-pipeline}
\end{figure*}

Figure~\ref{fig:asset-pipeline} shows the three-stage construction pipeline: 3D model creation, structural processing, and standardized preservation. Models are obtained from existing inventories or generated inside Pipette through Tencent Hunyuan, which supports text- and image-conditioned 3D asset generation as a built-in platform capability. The generated or imported models are then geometrically optimized. Rigid objects remain single bodies, whereas articulated instruments are decomposed, aligned, and reassembled according to their kinematics. Validated assets are stored in a standardized USDZ library.

The library currently contains over 100 assets. New objects can be generated, edited, and registered by providing standardized geometry, physics, collisions, and task semantics, then reused across scene construction, data collection, augmentation, training, and evaluation without modifying downstream pipelines.

\subsection{Language-Guided Scene Construction and Task Registration}\label{language-guided-scene-construction-and-task-registration}

\begin{figure}[!b]
\centering
\includegraphics[width=\columnwidth]{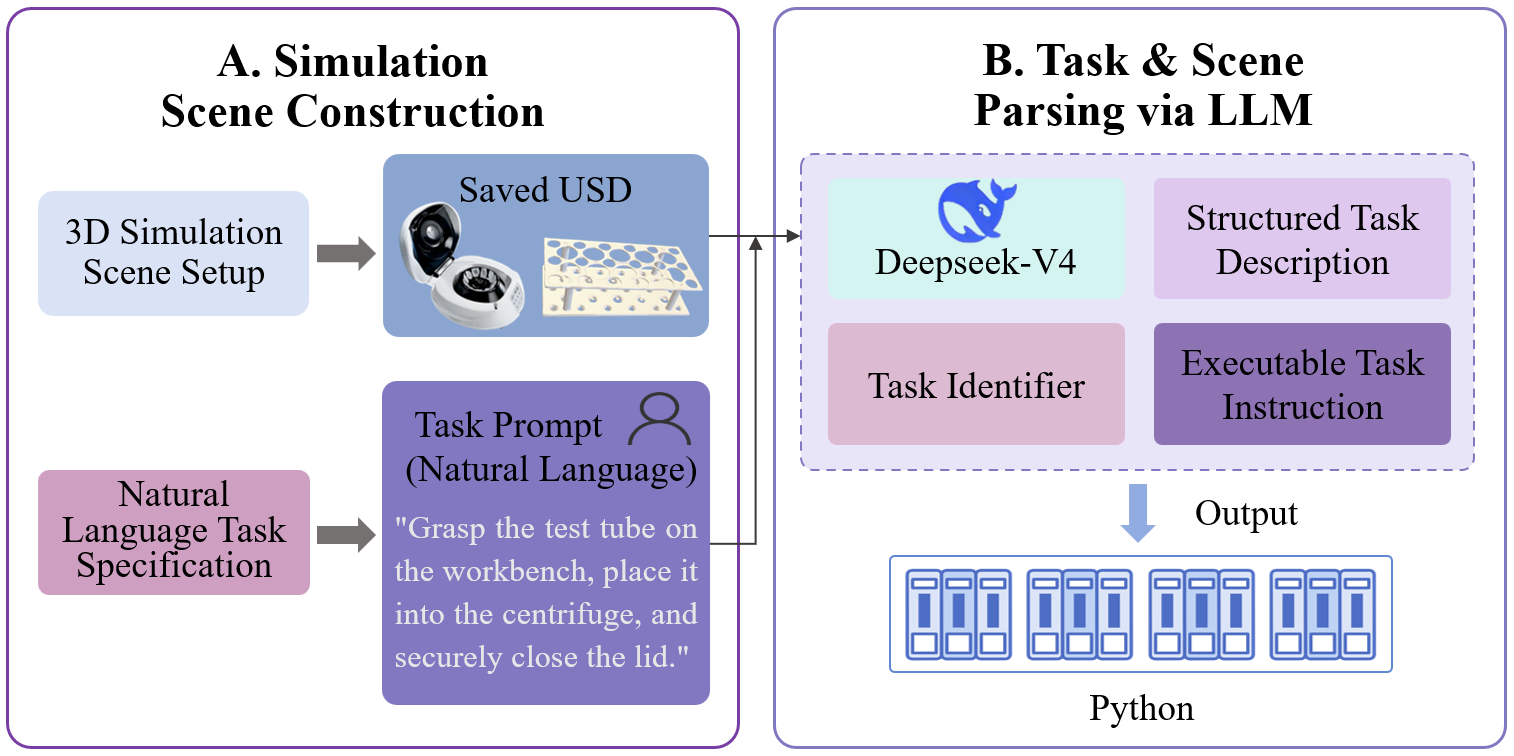}
\caption{Language-guided scene construction and task parsing workflow in Pipette Platform}
\label{fig:scene-construction}
\end{figure}

Pipette represents each wet-lab task as a structured entry containing the USD scene, robot initial state, sensor configuration, language instruction, target object, execution constraints, and success criterion. This representation separates task definition from data collection, augmentation, training, and evaluation, allowing tasks to share a common execution and evaluation interface.

Figure~\ref{fig:scene-construction} illustrates the registration workflow. A user builds and saves a 3D scene and describes the intended operation in natural language. An LLM-based parser extracts relevant objects, intent, and constraints, then drafts task identifiers, instructions, and structured Python configurations. At runtime, a shared initializer creates the Isaac Lab context, loads the configured robot (default: Franka Panda), restores its state, and configures the main, top, and wrist cameras. Reusing this initializer across demonstration collection, augmentation, and policy inference keeps scene, robot, and sensor conditions consistent. The parser assists task configuration rather than performing autonomous task planning. Environment-dependent fields, including prim and scene paths, camera settings, success thresholds, and evaluator logic, still require user verification through simulation replay and online evaluation. The mechanism therefore reduces per-task scripting while retaining explicit validation for reproducibility.

\subsection{Success-Verified Simulation Augmentation}\label{data-efficient-success-verified-simulation-augmentation}

Wet-lab policy learning is constrained by costly demonstrations, whereas image-space augmentation changes pixels without updating robot states, dynamics, or action labels. Pipette instead combines limited teleoperated demonstrations with simulation-level trajectory augmentation and task-level success verification. Replaying trajectories in physics simulation produces synchronized multi-view images, proprioceptive states, actions, timestamps, and success labels, preserving observation-action consistency. Demonstrations are collected through keyboard teleoperation, with damped least-squares inverse kinematics converting end-effector commands into robotic-arm joint targets. At each control step, Pipette records the three RGB views, proprioceptive and simulation states, action, and language instruction in HDF5. For augmentation, it restores and validates the original episode, then replays it under perturbations to lighting, execution speed, camera pose, and joint actions. Each replay is re-rendered and re-recorded rather than paired with the original observations, yielding physically consistent variations. Teleoperation and perturbation details are provided in the separately submitted technical supplement. Each replay is validated by a task evaluator that compares object poses, actuator states, or geometric relations with registered thresholds. The resulting success label can filter training episodes, while failed rollouts can support failure-mode analysis. Task-specific criteria are detailed in the separately submitted technical supplement.

\subsection{LeRobot Conversion, VLA Training and Evaluation}\label{lerobot-conversion-vla-training-and-evaluation}

Success-verified or manually annotated HDF5 episodes are converted to LeRobot format~\citep{cadene2026lerobot}. Each sample contains top-, main-, and wrist-view RGB images at 400 \ensuremath{\times} 400, a proprioceptive state comprising robotic-arm joints and the gripper state, a joint-target action with a gripper command, and a language instruction. The converter removes stale visual frames, aligns observations, states, and actions, and downsamples higher-frequency episodes to a unified 10 Hz rate~\citep{lan2025autobio}.

Raw and augmented episodes are converted separately. ACT, SmolVLA, and \(\pi\)0 use the same data interface, training entry, observation definition, and 8-dimensional robot-command interface. Comparisons hold the simulator, task definitions, observations, and success evaluators fixed while retaining each policy's native action decoding.

Online, all policies receive the common observations and language input and are evaluated in the shared simulator using task-specific evaluators. They execute 8-dimensional robot commands, while their native action decoding and query schedules are retained. Each episode is logged to JSON with its success status, failure reason, runtime, policy and control frequencies, gripper threshold, and evaluator metrics, supporting unified success-rate computation and failure analysis.

\section{Experiments}\label{experiments}

\subsection{Experimental Setup}\label{experimental-setup}

We evaluate ACT, SmolVLA, and \ensuremath{\pi}0 in a unified Isaac Sim/Isaac Lab setup with shared observations, an 8D robot-command interface, task definitions, and success criteria; policy-specific action decoding is retained. Each policy is trained with raw demonstrations and with simulation-augmented data, and is evaluated by task success rate. Full environment settings, data configurations, and model-specific training parameters are provided in the separately submitted technical supplement.

\subsection{Wet-Lab Embodied Task Benchmark}\label{wet-lab-embodied-task-benchmark}

The benchmark contains 12 representative wet-lab manipulation tasks spanning sample and culture-consumable handling, equipment-lid control, and instrument placement or relocation. Together, these tasks assess visual localization, pose control, target alignment, contact stability, and long-horizon execution. The complete task taxonomy and associated competencies are detailed in the separately submitted technical supplement.

\subsection{Policy Training and Evaluation on Pipette}\label{policy-training-and-evaluation-on-pipette}

Simulation augmentation has strongly category-dependent effects (Table~\ref{tab:category-success}). For sample and culture tasks, average success increased from 59.5\% to 69.3\% for ACT, 36.3\% to 90.0\% for SmolVLA, and 41.8\% to 53.3\% for \(\pi\)0. SmolVLA reached at least 86\% on all four tasks and recovered petri-dish retrieval from 0\% to 92\%, indicating that augmentation effectively expands object appearance and target-pose coverage.

\begin{table}[t]
\centering
\begingroup
\small
\begin{tabular}{@{}lcccc@{}}
\toprule
\textbf{Task Category} & \textbf{Setting} & \textbf{ACT} & \textbf{SmolVLA} & \(\pi\)\textbf{0} \\
\midrule
Sample/culture & Unenh. & 59.5 & 36.3 & 41.8 \\
& Enh. & \textbf{69.3} & \textbf{90.0} & \textbf{53.3} \\
Hatch control & Unenh. & 79.3 & 63.0 & 65.8 \\
& Enh. & \textbf{83.5} & \textbf{83.5} & \textbf{68.0} \\
Placement/relocation & Unenh. & \textbf{42.3} & 22.0 & 4.3 \\
& Enh. & 24.3 & \textbf{42.0} & \textbf{11.0} \\
\bottomrule
\end{tabular}
\endgroup
\caption{Success rates by task category}
\label{tab:category-success}
\end{table}

\begin{table*}[t]
\centering
\begingroup
\small
\begin{tabular}{@{}
  >{\raggedright\arraybackslash}m{(\linewidth - 10\tabcolsep) * \real{0.17}}
  >{\raggedright\arraybackslash}m{(\linewidth - 10\tabcolsep) * \real{0.33}}
  >{\centering\arraybackslash}m{(\linewidth - 10\tabcolsep) * \real{0.13}}
  >{\centering\arraybackslash}m{(\linewidth - 10\tabcolsep) * \real{0.10}}
  >{\centering\arraybackslash}m{(\linewidth - 10\tabcolsep) * \real{0.15}}
  >{\centering\arraybackslash}m{(\linewidth - 10\tabcolsep) * \real{0.12}}@{}}
\toprule
\textbf{Task Category} & \textbf{Specific Task} & \textbf{Setting} & \textbf{ACT} & \textbf{SmolVLA} & \(\pi\)\textbf{0} \\
\midrule
\multirow{8}{=}{Sample and culture consumables handling tasks}
& \multirow{2}{=}{Pick up test tube} & Unenh. & 100 & 48 & 72 \\
& & Enh. & 100 & \textbf{86} & \textbf{87} \\
& \multirow{2}{=}{Pipette-to-petri-dish positioning} & Unenh. & \textbf{81} & 41 & 32 \\
& & Enh. & 78 & \textbf{90} & \textbf{66} \\
& \multirow{2}{=}{Remove the petri dish from the incubator} & Unenh. & 0 & 0 & 29 \\
& & Enh. & \textbf{9} & \textbf{92} & \textbf{30} \\
& \multirow{2}{=}{Place the petri dish in the incubator} & Unenh. & 57 & 56 & \textbf{34} \\
& & Enh. & \textbf{90} & \textbf{92} & 30 \\
\midrule
\multirow{8}{=}{Equipment hatch and opening/closing control tasks}
& \multirow{2}{=}{Close the centrifuge lid} & Unenh. & 76 & 6 & 23 \\
& & Enh. & \textbf{84} & \textbf{83} & \textbf{52} \\
& \multirow{2}{=}{Open the centrifuge lid} & Unenh. & 88 & 91 & \textbf{92} \\
& & Enh. & \textbf{100} & \textbf{99} & 87 \\
& \multirow{2}{=}{Open the water bath lid} & Unenh. & 70 & \textbf{100} & \textbf{51} \\
& & Enh. & \textbf{98} & 85 & 33 \\
& \multirow{2}{=}{Close the spectrophotometer lid} & Unenh. & \textbf{83} & 55 & 97 \\
& & Enh. & 52 & \textbf{67} & \textbf{100} \\
\midrule
\multirow{8}{=}{Instrument placement and relocation tasks}
& \multirow{2}{=}{Place the centrifuge tube on the electronic balance} & Unenh. & \textbf{58} & 22 & 2 \\
& & Enh. & 3 & \textbf{27} & \textbf{6} \\
& \multirow{2}{=}{Remove the centrifuge tube from the electronic balance} & Unenh. & 40 & 56 & 12 \\
& & Enh. & \textbf{71} & \textbf{60} & 12 \\
& \multirow{2}{=}{Place the pipette in the pipette rack} & Unenh. & \textbf{67} & 10 & 0 \\
& & Enh. & 5 & \textbf{41} & \textbf{9} \\
& \multirow{2}{=}{Place the Erlenmeyer flask on the shaker} & Unenh. & 4 & 0 & 3 \\
& & Enh. & \textbf{18} & \textbf{40} & \textbf{17} \\
\bottomrule
\end{tabular}
\endgroup
\caption{Task-level success rates across the 12 wet-lab benchmark tasks. Each task is evaluated with unenhanced demonstrations (Unenh.) and simulation-augmented demonstrations (Enh.).}
\label{tab:task-success}
\end{table*}

Hatch control was the most stable category: ACT, SmolVLA, and \(\pi\)0 improved from 79.3\%, 63.0\%, and 65.8\% to 83.5\%, 83.5\%, and 68.0\%, respectively. The largest gain was SmolVLA on centrifuge-lid closing, from 6\% to 83\%. These tasks involve relatively fixed components and success conditions, making their pose and contact distributions easier to preserve during replay.

Placement and transfer remained most difficult. When the task in which an AgileX robotic arm places an Erlenmeyer flask on a shaker is included, augmentation raised SmolVLA from 22.0\% to 42.0\% and \(\pi\)0 from 4.3\% to 11.0\%, but reduced ACT from 42.3\% to 24.3\%. For ACT, balance placement fell from 58\% to 3\% and rack placement from 67\% to 5\%, although balance removal improved from 40\% to 71\% and the success rate for placing the Erlenmeyer flask on the shaker improved from 4\% to 18\%. Thus, broad pose or contact perturbations can weaken behavior-cloning policies on tasks requiring low-height grasping, precise release, and stable contact. Detailed task-level results are reported in Table~\ref{tab:task-success}.

Under the fixed 30-demonstration budget, augmentation often improves success but is not uniformly beneficial: \(\pi\)0 remained below 20\% on all placement tasks, and ACT degraded on two precision-placement tasks. Performance depends jointly on model architecture, action representation, contact accuracy, robotic embodiment, and whether perturbations preserve task-relevant pose and action distributions.

\subsection{Failure Analysis and Case Study}\label{failure-analysis-and-case-study}

\begin{figure}[!b]
\centering
\includegraphics[width=0.72\columnwidth]{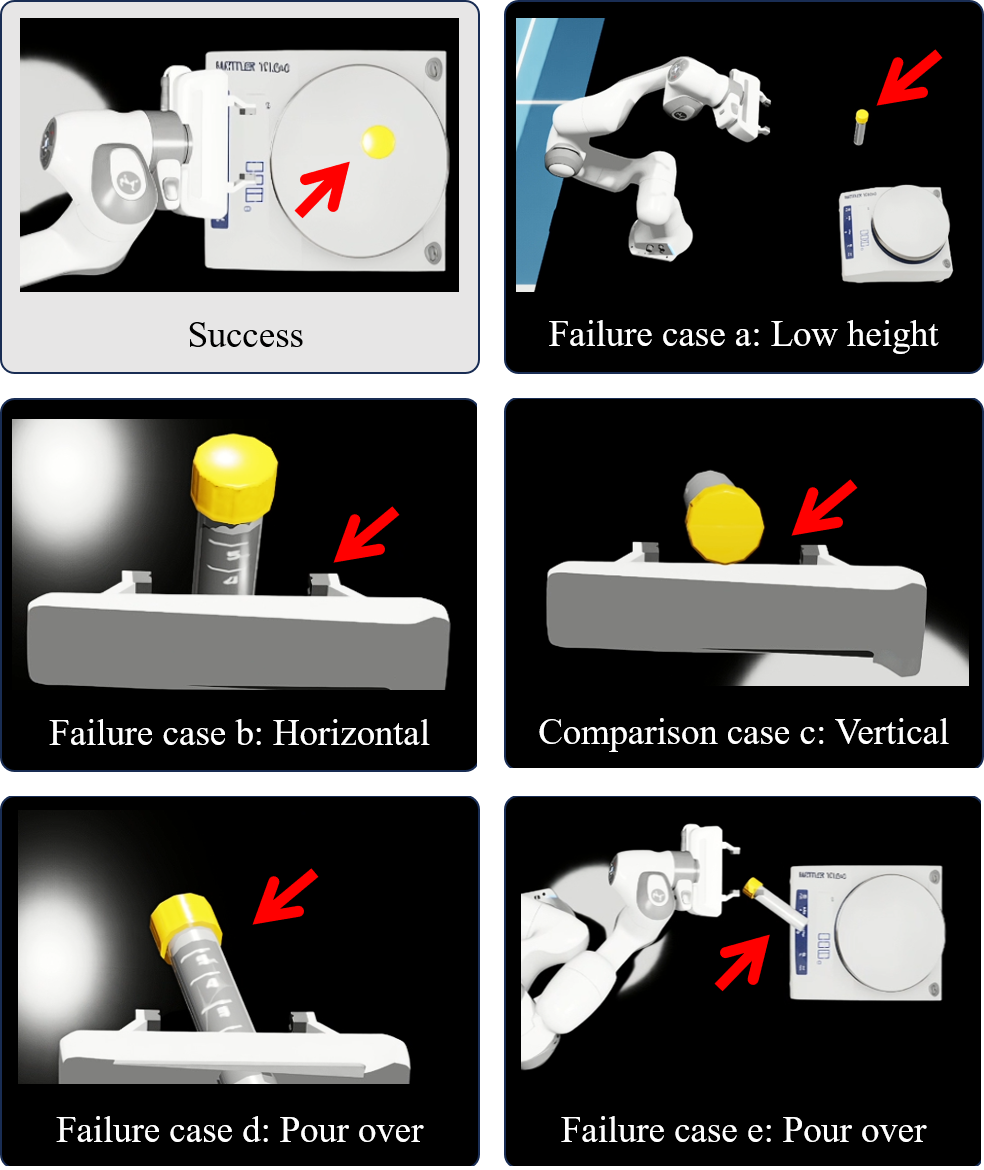}
\caption{Success and failure cases for balance placement}
\label{fig:failure-cases}
\end{figure}

Figure~\ref{fig:failure-cases} shows success and representative failures in placing a centrifuge tube on an electronic balance. Success requires the tube to remain in the balance target region and height range for the hold time. The low, free-standing tube, lateral grasp, and precise placement make small pose, timing, or contact errors tip or displace it.

Because the tube starts near the table, the robot approaches laterally near the workspace boundary (Figure~\ref{fig:failure-cases}(a)). We use the lateral grasp in Figure~\ref{fig:failure-cases}(b), rather than the vertical grasp in Figure~\ref{fig:failure-cases}(c), because wet-lab workflows often continue with pouring or transfer, where side grasping is more appropriate. This choice, however, narrows the clearance to the table and tube wall. Inaccurate approach poses can push the tube before closure, causing tipping, rolling, or displacement (Figure~\ref{fig:failure-cases}(d)); even after a successful grasp, release-angle errors can leave the tube outside the effective balance area (Figure~\ref{fig:failure-cases}(e)).

This case shows that wet-lab manipulation failures reflect not only target localization, but also task semantics, object pose, and contact stability. Low, small, and unsecured consumables require more precise pose prediction, contact-aware control, and recovery.

\section{Conclusion}\label{conclusion}

This paper presents Pipette, a wet-lab embodied simulation platform that integrates scalable asset construction, language-driven task registration, success-verified simulation augmentation, and unified VLA training and evaluation. Across 12 representative manipulation tasks, Pipette enables controlled evaluation of ACT, SmolVLA, and \ensuremath{\pi}0 under common observation modalities, task definitions, and success criteria while retaining policy-specific action decoding. With 30 manual demonstrations, validated augmentation often improves success, especially for sample handling and equipment operation, while failure cases expose challenges in small-object grasping, low-height manipulation, precise placement, and contact stability.

Key limitations and future directions:

\textbf{Single-arm manipulation.} Pipette is currently evaluated mainly with single-arm simulated manipulation. Future work will extend the platform to dexterous hands and dual-arm systems for more complex wet-lab operations.

\textbf{Sim-to-real validation.} Current experiments are conducted in simulation. Real-robot deployment under sensing, lighting, material, and contact uncertainty remains an important next step.

\textbf{Automated evaluation.} Pipette now relies on code-based success evaluators. VLM-based task evaluation could reduce manual rule design and improve scalability for wet-lab tasks.

\bibliography{references}

% Technical appendix content merged from appendix.tex for the arXiv preprint.
\maketitle

\appendix

\section{More Information about the Pipette Platform}\label{a-more-information-about-the-pipette-platform}

\subsection{Introduction to USD Asset Structures}\label{a.1-introduction-to-usd-asset-structures}

USD (Universal Scene Description) assets are an open 3D scene and asset description format proposed by Pixar, commonly used to build complex digital scenes, robotic simulation environments, and digital twin systems. USD organizes scene content through a hierarchical structure, unifying information such as geometric models, materials, textures, lighting, cameras, physical properties, colliders, joint constraints, and semantic tags into components such as Stage, Prim, Attribute, and Relationship. For wet lab simulation platforms, USD assets are used not only to render the appearance of laboratory instruments and consumables but also to define objects' interactive properties, enabling robotic arms to grasp, place, or press objects such as centrifuge tubes, pipettes, petri dishes, centrifuge lids, and electronic balances, or to read them as targets for task completion verification. Leveraging USD's non-destructive editing and composition capabilities, the platform can extend new laboratory objects and task scenarios without modifying the underlying simulation workflow.

USDZ is a packaging format for USD assets, typically understood as a portable asset package that bundles USD scene files and their associated resources. Compared to standard USD files, USDZ is better suited for asset distribution, cross-platform viewing, and lightweight rendering, as it centralizes dependent resources such as models, materials, and textures, thereby reducing issues related to missing paths or incorrect resource references. During the construction of simulation platforms, USDZ assets can serve as the initial source or intermediate format for 3D objects. After undergoing scale calibration, coordinate system alignment, collision body generation, physical property configuration, and task semantic annotation, they can be converted or integrated into standardized USD assets.

\subsection{Instrument Assets}\label{a.2-instrument-assets}

\begin{figure*}[t]
\centering
\setlength{\tabcolsep}{2pt}
\begin{tabular}{ccc}
\includegraphics[width=0.29\textwidth,height=0.13\textheight,keepaspectratio]{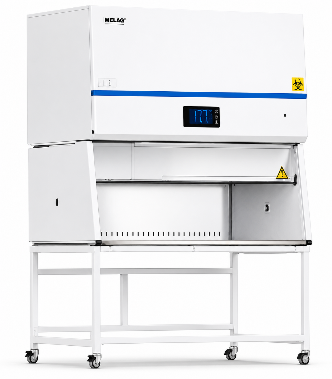} & \includegraphics[width=0.29\textwidth,height=0.13\textheight,keepaspectratio]{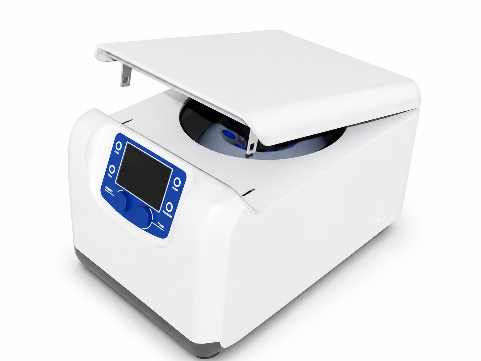} & \includegraphics[width=0.29\textwidth,height=0.13\textheight,keepaspectratio]{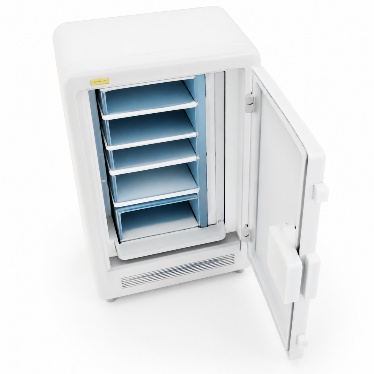} \\
\includegraphics[width=0.29\textwidth,height=0.13\textheight,keepaspectratio]{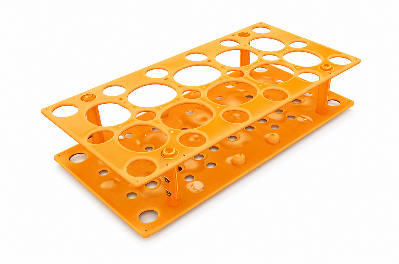} & \includegraphics[width=0.29\textwidth,height=0.13\textheight,keepaspectratio]{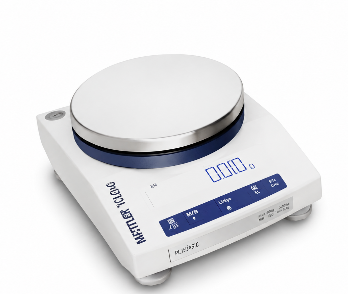} & \includegraphics[width=0.29\textwidth,height=0.13\textheight,keepaspectratio]{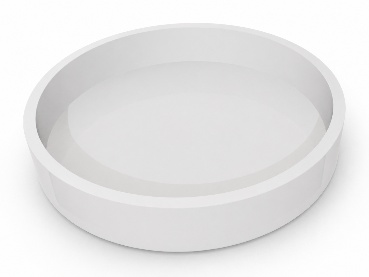} \\
\includegraphics[width=0.29\textwidth,height=0.13\textheight,keepaspectratio]{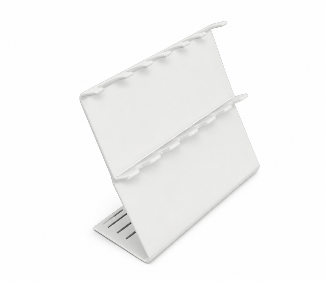} & \includegraphics[width=0.29\textwidth,height=0.13\textheight,keepaspectratio]{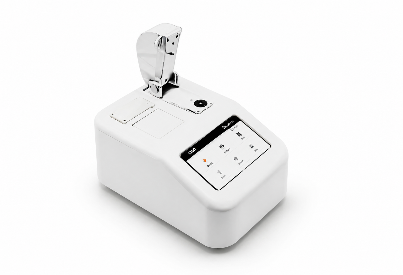} & \includegraphics[width=0.29\textwidth,height=0.13\textheight,keepaspectratio]{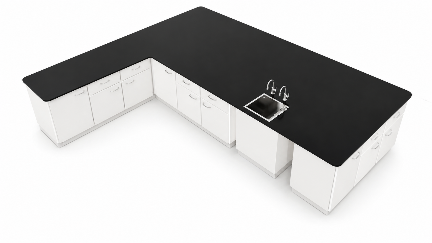} \\
\includegraphics[width=0.29\textwidth,height=0.13\textheight,keepaspectratio]{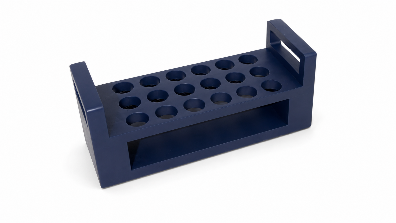} & \includegraphics[width=0.29\textwidth,height=0.13\textheight,keepaspectratio]{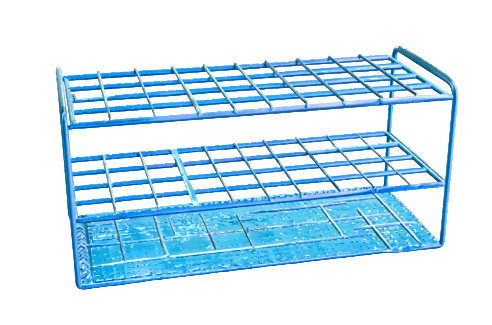} & \includegraphics[width=0.29\textwidth,height=0.13\textheight,keepaspectratio]{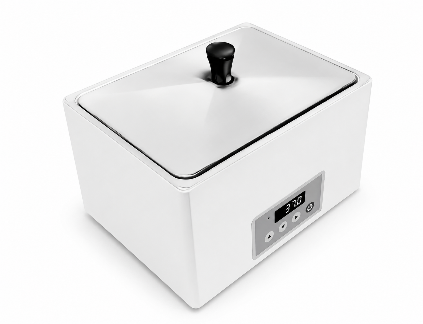} \\
\end{tabular}
\caption{Representative interactive wet-lab assets included in the Pipette platform (part 1).}
\label{fig:wet-lab-assets-1}
\end{figure*}

\begin{figure*}[t]
\centering
\setlength{\tabcolsep}{2pt}
\begin{tabular}{ccc}
\includegraphics[width=0.29\textwidth,height=0.13\textheight,keepaspectratio]{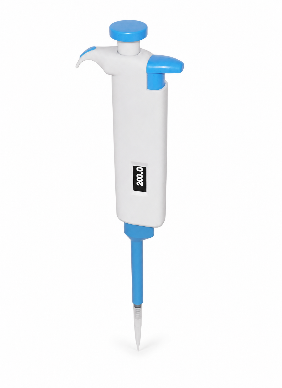} & \includegraphics[width=0.29\textwidth,height=0.13\textheight,keepaspectratio]{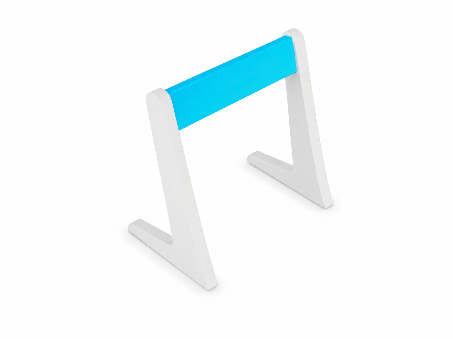} & \includegraphics[width=0.29\textwidth,height=0.13\textheight,keepaspectratio]{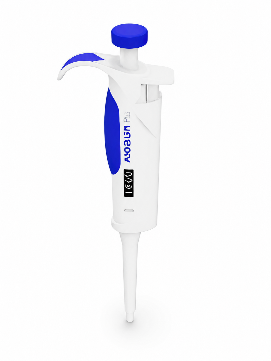} \\
\includegraphics[width=0.29\textwidth,height=0.13\textheight,keepaspectratio]{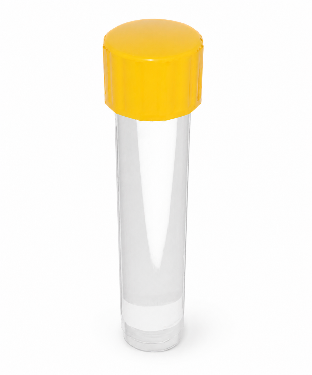} & \includegraphics[width=0.29\textwidth,height=0.13\textheight,keepaspectratio]{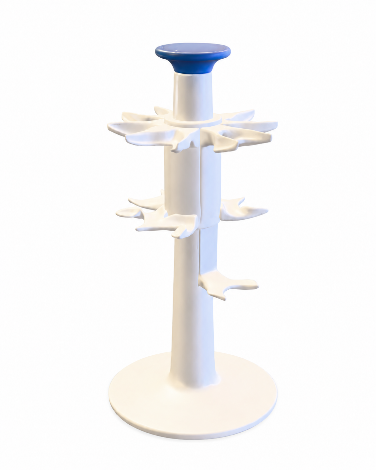} & \includegraphics[width=0.29\textwidth,height=0.13\textheight,keepaspectratio]{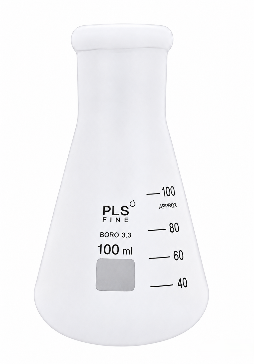} \\
\includegraphics[width=0.29\textwidth,height=0.13\textheight,keepaspectratio]{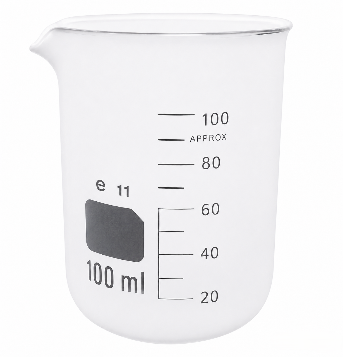} & \includegraphics[width=0.29\textwidth,height=0.13\textheight,keepaspectratio]{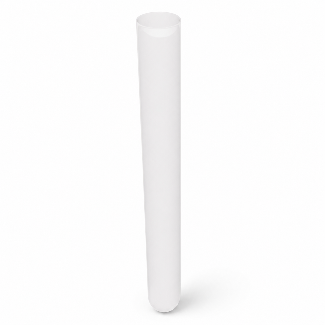} & \includegraphics[width=0.29\textwidth,height=0.13\textheight,keepaspectratio]{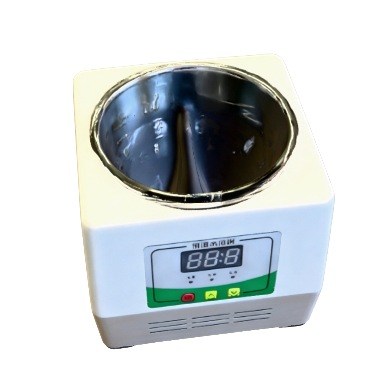} \\
\end{tabular}
\caption{Representative interactive wet-lab assets included in the Pipette platform (part 2).}
\label{fig:wet-lab-assets-2}
\end{figure*}

Figures~\ref{fig:wet-lab-assets-1} and~\ref{fig:wet-lab-assets-2} present representative interactive instrument and consumable assets included in the Pipette platform. The two-part overview groups the available wet-lab equipment into a compact visual inventory while preserving the individual appearance of each asset for reference.

\subsection{Simulation Timing and Observation-Action Logging}\label{a.3-simulation-timing-and-observation-action-logging}

Set the simulation time step to

\begin{equation}
\Delta t_{sim} = \frac{1}{60}\, s,
\end{equation}

The default closed-loop control frequency is\(f_{c} = 30Hz\), and the vision sampling and policy query frequency is\(f_{v} = f_{\pi} = 10Hz\). Under the default parameters, control commands are executed every 2 simulation steps, and new vision frames are acquired every 3 control steps. At each control step, three 400\ensuremath{\times}400 RGB images, the robot arm's body state, actions, simulation state, and voice commands are recorded. The state and actions are written as\\
\begin{equation}
\begin{aligned}s_{t} &= \left\lbrack q_{1,t},\ldots,q_{7,t},g_{t} \right\rbrack \in \mathbb{R}^{8},\\a_{t} &= \left\lbrack q_{1,t}^{\star},\ldots,q_{7,t}^{\star},u_{g,t} \right\rbrack \in \mathbb{R}^{8}.\end{aligned}
\end{equation}

Here, \(q_{i}\)represents the positions of Franka's 7 joints, \(g_{t}\) represents the joint state of the single-sided gripper, \(q_{i}^{\star}\) represents the target joint positions, and \(u_{g} \in \left\{ 0,1 \right\}\) denotes the gripper's open/close command. Specifically, \(u_{g} = 1\) indicates the gripper is open, and \(u_{g} = 0\ \)indicates the gripper is closed. If there is no new image for the current control step, reuse the most recent vision frame and write it to

\begin{equation}
\begin{aligned}{fresh}_{t} &= \mathbf{1}\left\{(t-1)\,{mod}\,d_{v}=0\right\},\\{age}_{t} &= t-t_{last\_fresh}.\end{aligned}
\end{equation}
The timestamp records both the simulation time \(t_{sim} = n_{sim}\Delta t_{sim}\) and the wall-clock time to enable accurate playback and time-scaling augmentation later.

\subsection{Policy Communication Transmission}\label{a.4-policy-communication-transmission}

The online evaluation uses ZeroMQ REQ/REP communication. Inference clients connect to the default endpoint `tcp://127.0.0.1:5555` and send a Python object dictionary to the server. The main fields are `observation.images.top`, `observation.images.main`, `observation.images.wrist`, `observation.state`, `task` or `language\_instruction`, and `reset\_policy`. The ACT client sends CHW images normalized to {[}0,1{]} as float32; the SmolVLA and \ensuremath{\pi}0 clients send CHW images as uint8. Before feeding them into the model, the server converts the uint8 images to float32 and divides them by 255.

The server loads the corresponding LeRobot policy, performs preprocessing, `select\_action`, and postprocessing, and returns a single-step action \(a_{t} \in \mathbb{R}^{8}\) or an action block\(\ A_{t} \in \mathbb{R}^{K \times 8}\). The communication timeout is 65,000 ms for the ACT and SmolVLA clients, and 300,000 ms for the \ensuremath{\pi}0 client. If consecutive communication failures reach the threshold, the client enters a safe hold state:

\begin{equation}
a_{hold} = \left\lbrack q_{now},1 \right\rbrack,
\end{equation}

Specifically, the robotic arm maintains its current 7-joint position, and the gripper remains open. The consecutive failure threshold for ACT and SmolVLA is 6 times, and for \ensuremath{\pi}0 it is 20 times. After all episodes have ended, the system calculates

\begin{equation}
SR = \frac{N_{success}}{N_{success} + N_{failure}},
\end{equation}

Write the success status, failure reason, runtime, control frequency, policy frequency, gripper thresholds, and evaluator metrics to a JSON file.

\subsection{Teleoperation Parameters and Inverse Kinematics}\label{a.5-teleoperation-parameters-and-inverse-kinematics}

Teach-in data collection uses the SE(3) keyboard remote control interface. The default sensitivity coefficient in the task preset is k=4.0; the code converts this to

\begin{equation}
k_{p} = 0.002k = 0.008,\ \ k_{r} = 0.01k = 0.04,
\end{equation}

Used for translation and rotation inputs, respectively. At each control step, the system reads the incremental keyboard inputs \ensuremath{\Delta}p and \ensuremath{\Delta}\ensuremath{\phi} and updates the target pose of the end-effector:

\begin{equation}
p_{t + 1}^{\star} = p_{t}^{\star} + \Delta p,\ q_{t + 1}^{\star} = normalize\left( q_{t}^{\star} \otimes Quat(\Delta\phi) \right).
\end{equation}

The Differential IK controller then uses damped least squares to transform the end-effector targets into 7-dimensional joint targets. Its equivalent form can be written as

\begin{equation}
\Delta q = J^{\top}\left( JJ^{\top} + \lambda^{2}I \right)^{- 1}e,
\end{equation}

Here, J is the end-effector Jacobian matrix, e is the end-effector pose error, and the damping coefficient \ensuremath{\lambda} = 0.05. The target position of the gripper is controlled using binary control: when open, the target position of both fingers is 0.04 m, and when closed, it is 0. To minimize object slippage, the stiffness of the gripper actuator is set to 10\textsuperscript{4}, the damping to 10\textsuperscript{3}, and the effort limit to 1000 in the simulation.

\subsection{Online Motion Execution, Smoothing, and Safe Holding}\label{a.6-online-motion-execution-smoothing-and-safe-holding}

In the online evaluation, the control frequency is 30 Hz, and the policy is typically queried at 10 Hz, so

\begin{equation}
d_{\pi} = \max\left( 1,round\left( \frac{f_{c}}{f_{\pi}} \right) \right) = 3.
\end{equation}

The ACT client performs linear interpolation on the robot joint targets between two policy queries:

\begin{equation}
\alpha = \min\left( \frac{m}{d_{\pi}},1 \right),\ \ q^{cmd} = (1 - \alpha)q_{prev} + \alpha q_{curr},
\end{equation}

Here, \ensuremath{m} is the number of control steps already executed under the current policy goal. The gripper dimensions are not interpolated; instead, the gripper is determined to be open using the threshold \(u_{g} \geq 0.5\). If SmolVLA and \ensuremath{\pi}0 return an action block \(A \in \mathbb{R}^{K \times 8}\), they execute \(A_{0},A_{1},\ldots,A_{K - 1}\) sequentially in the control loop until the next policy query overrides the action block. SmolVLA uses a default gripper threshold of 0.5; \ensuremath{\pi}0 employs a hysteresis threshold, opening the gripper only when\(u_{g} \geq 0.55\) while it is currently closed, and maintaining the open state only when \(u_{g} > 0.45\) while it is currently open, thereby reducing opening and closing oscillations near the threshold.

The centrifuge lid-opening task utilizes an additional virtual button mechanism. When the gripper's sensing point enters the button's trigger zone, the target angle of the lid's revolute joint linearly changes to -40\ensuremath{^{\circ}} within 2 seconds, with drive parameters set to max force 1000, stiffness 200, and damping 20. The lid-opening task is considered successful if the button has been triggered and the minimum target angle is less than -39\ensuremath{^{\circ}}.

\subsection{Simulated Data Augmentation}\label{a.7-simulated-data-augmentation}

The augmentation script does not directly modify the offline image; instead, it re-executes the original motion path in Isaac Sim. Lighting augmentation scales the intensity of the light sources visible in the scene proportionally:

\begin{equation}
I' = s_{I}I,
\end{equation}

The default value for \(s_{I} = 0.8\). Time scaling uses the speed factor \(s_{v}\), with a default value of \(s_{v} = 1.2\). If you only want to scale existing timestamps, use

\begin{equation}
t'_{i} = t_{0} + \frac{t_{i} - t_{0}}{s_{v}}.
\end{equation}

When resampling, the script first generates a sequence of source track positions

\begin{equation}
x_{0} = 0,\ \ x_{i} = x_{i - 1} + s_{v}m_{i},
\end{equation}

where \(m_{i} = clip\left( 1 + \rho\epsilon_{i},0.2,3.0 \right)\), \(\rho = 0.10\), and \(\epsilon_{i}\) is a local velocity perturbation obtained by smoothing Gaussian noise with a sliding average of length 11 and then normalizing it. Motion interpolation uses Catmull-Rom cubic splines by default:

\begin{equation}
\begin{aligned}a(x)=\frac{1}{2}\bigl(&2p_{1}+(-p_{0}+p_{2})t\\&+(2p_{0}-5p_{1}+4p_{2}-p_{3})t^{2}\\&+(-p_{0}+3p_{1}-3p_{2}+p_{3})t^{3}\bigr).\end{aligned}
\end{equation}

Here, \(p_{0},p_{1},p_{2},p_{3}\) are adjacent frames, and \(t = x - \left\lfloor x \right\rfloor\). You can also switch to linear interpolation.

Bounded Gaussian noise superimposed on the target positions of the first 7 joints of the robotic arm:

\begin{equation}
\eta\mathcal{\sim N}\left( 0,{0.005}^{2} \right),\ \ \eta \leftarrow clip(\eta, - 0.02,0.02)\ rad.
\end{equation}

Uniform sampling of camera motion within a 3D sphere:

\begin{equation}
\delta = rU^{1/3}\frac{n}{\left\| n \right\|},\ \ U\mathcal{\sim U}(0,1),\ \ n\mathcal{\sim N}\left( 0,I_{3} \right).
\end{equation}

The base radius recorded in the script is 0.01 m; however, during actual sampling, the following coverage radii were used for the cameras: 0.002 m for the wrist camera and 0.2 m for the other cameras. The enhanced output is compressed using gzip level 4, with an RGB chunk shape of 1, H, W, C. All enhanced variants are re-evaluated; variants that fail validation are discarded, and only successfully enhanced samples are retained in the output file.

\subsection{Task Success Criteria}\label{a.8-task-success-criteria}

Most task success criteria use continuous hold time as the final condition. The test tube grasping task requires that the height meet the condition \(z_{t} > z_{0} + \Delta z\ \)and be maintained for a specified duration, while the tilt relative to the initial orientation:

\begin{equation}
\theta = \arccos\left( {\widehat{u}}_{t}^{\top}{\widehat{u}}_{0} \right)
\end{equation}

If the failure threshold is exceeded, the task fails. By default, the test tube pickup task uses \(\Delta z = 0.02\, m\), a hold time of 0.5 s, a tilt failure threshold of 35\ensuremath{^{\circ}}, and a timeout of 7 s; The task for removing the centrifuge tube from the electronic balance uses \(y_{t} - y_{0} > 0.25\, m\), \(z_{0} - z_{t} > 0.12\, m\), a hold time of 1.2 s, a tilt failure threshold of 10\ensuremath{^{\circ}}, and a timeout of 10 s.

Use the XY distance when the pipette picks up or deposits a sample in the target area

\begin{equation}
d_{xy} = \left\| p_{pipette}^{xy} - p_{target}^{xy} \right\|_{2}
\end{equation}

The primary success criteria are \(d_{xy} \leq 0.12\, m\). If \(z < 0.6\, m\), the task is deemed a failure due to insufficient height, and the time limit is 12 seconds. The pipette rack placement task further requires \(y > 0.14\, m\) and \(z > 0.8\, m\), with these conditions maintained for 1 second.

For lid-related tasks, angle or displacement thresholds are used. For the centrifuge lid opening, the lid's angle in the x-direction must be greater than -46\ensuremath{^{\circ}} and maintained for 0.25 s; for the spectrophotometer lid opening, \(x > 90^{\circ}\) and maintained for 0.5 s; For the water bath, opening the lid requires \(x_{0} - x_{t} > 0.30\, m\) and \(z_{t} - z_{0} > 0.10\, m\), maintained for 0.3 s.

The petri dish retrieval task requires \(x < 0.6\, m\) and \(z > 0.6\, m\), and must be maintained for 0.5 s; The petri dish placement task requires \(x > 0.55\, m\), \(y < 0.1\, m\), and \(z > 0.6\, m\), with the height change between consecutive control steps \(\left| z_{t} - z_{t - 1} \right| \leq 0.002\, m\) and stability maintained for 0.5 s. For the centrifuge tube placement on the electronic balance task, project the balance pan onto an XY circular region with a radius equal to half the shorter diameter of the pan's bounding box multiplied by 0.5; the success condition is that the centrifuge tube's XY position enters this circular region, and \(0.63 < z < 0.73\, m\) is maintained for 1 s.

\section{About the Agent Feature in Pipette Platform}\label{b-about-the-agent-feature-in-pipette-platform}

To lower the barrier to entry for the Pipette Platform, this paper introduces a natural language agent module designed to assist users in completing processes such as task registration, data collection, data transformation, model training, and online inference. This module does not alter the data formats or policy model structures of Isaac Sim, Isaac Lab, or LeRobot; rather, it serves as a high-level interaction interface that translates users' natural language commands into structured task configurations and executable script commands.

The Agent primarily handles intent recognition, parameter extraction, and command generation. For commands related to data collection, format conversion, model training, or online evaluation, the Agent determines the corresponding workflow and populates parameters such as task name, data path, model type, output directory, and number of training steps based on the task registry and script directory. If information is incomplete, the Agent will prompt for the missing details; if the information is complete, it will generate the command and initiate the corresponding workflow.

During task construction, the Agent can convert wet lab operation descriptions into structured task entries that include scene configuration, robot initial state, camera settings, voice commands, target objects, and success criteria. It should be noted that the Agent currently primarily serves to coordinate workflows and assist with configuration; users must still verify key fields in the generated configuration, such as the target object's prim path, USD scene path, and success threshold. For complex tasks, task executability and evaluator validity still require verification through simulation playback and online evaluation. Therefore, this paper treats the Agent as a platform support module rather than an independent policy learning model or an automated task planning system.

\section{Task Details for the Pipette Platform Benchmark}\label{c-task-details-for-the-pipette-platform-benchmark}

\subsection{Sample and Culture Consumables Handling Tasks}\label{c.1-sample-and-culture-consumables-handling-tasks}

\subsubsection{Pick Up the Test Tube}\label{c.1.1-pick-up-the-test-tube}

\begin{figure*}[t]
\centering
\includegraphics[width=0.95\textwidth]{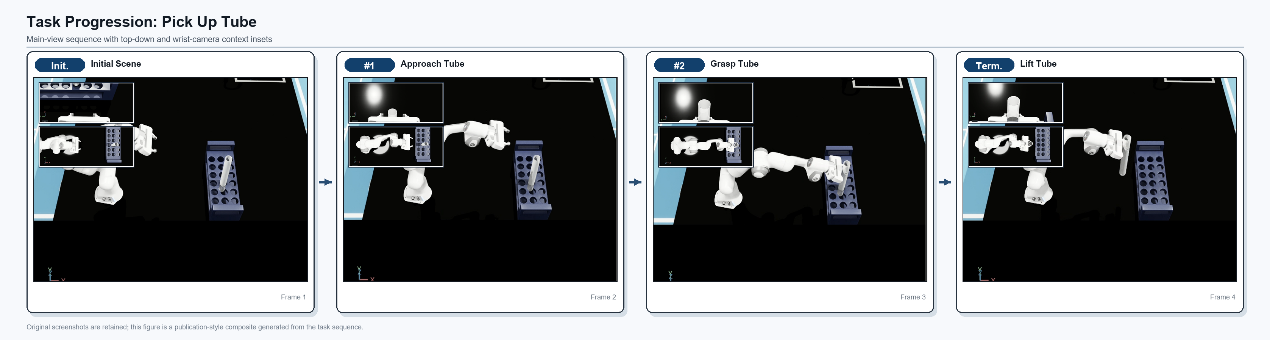}
\caption{Task sequence for Pick Up the Test Tube.}
\label{fig:task-27}
\end{figure*}

\par\noindent
\textbf{Description.} This task requires the robotic arm to pick up a test tube from the laboratory bench. At the start of the task, the Franka Panda robotic arm is positioned above the test tube. The policy must determine the test tube's location based on multi-view RGB images and the robotic arm's state. The execution process includes: first, the end-effector moves to a position above the test tube; then, it descends to an appropriate grasping height; next, the gripper's orientation is adjusted to align with the test tube's axis; then, the gripper closes to secure the test tube; finally, the test tube is lifted upward and held steadily.\par
\noindent\textbf{Success Criteria.} The test tube must be securely gripped by the gripper and must not be knocked over or tilt significantly during the gripping process. Once gripping is complete, the test tube must be lifted to a sufficient height relative to its initial position and remain stable for the specified duration. If the test tube is successfully lifted, does not fall, and does not exceed the tilt failure threshold, the task is deemed successful.\par

\subsubsection{Pipette-to-Petri-Dish Positioning}\label{c.1.2-pipette-to-petri-dish-positioning}

\begin{figure*}[t]
\centering
\includegraphics[width=0.95\textwidth]{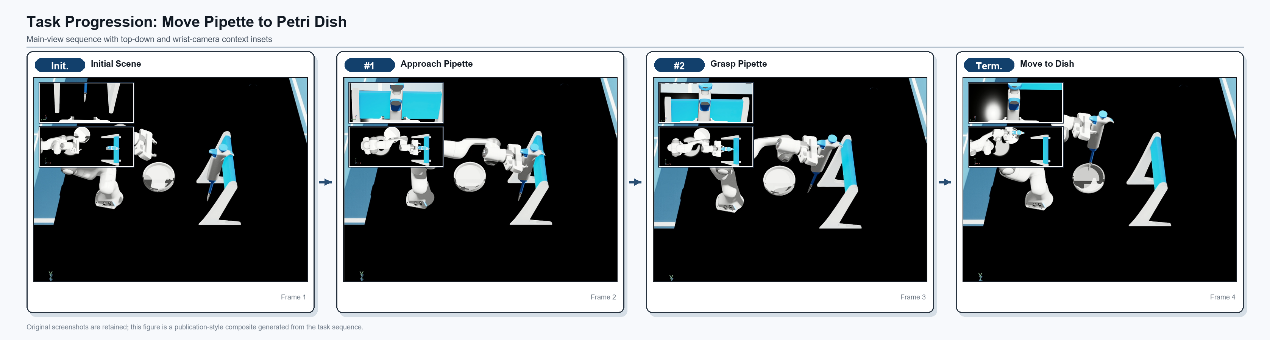}
\caption{Task sequence for Pipette-to-Petri-Dish Positioning.}
\label{fig:task-28}
\end{figure*}

\par\noindent
\textbf{Description.} This task requires the robotic arm to pick up the pipette from the lab bench and move it near the petri dish. At the start of the task, the pipette and petri dish are located on the lab bench. The policy must identify the pipette's slender structure and the target area for the petri dish. The execution process includes: the robotic arm moving above the pipette, adjusting the end-effector orientation to align with the pipette's grasping direction, descending and closing the grippers to complete the grasp, and then lifting the pipette and moving it to the target area near the petri dish.\par
\noindent\textbf{Success Criteria.} The pipette must be securely gripped and moved to a position near the target area on the petri dish. The task is considered successful if the horizontal distance between the tip or body of the pipette and the petri dish is less than the set threshold, and the pipette does not drop or fall below the safety height. If the pipette reaches the target area, the task is deemed successful.\par

\subsubsection{Remove the Petri Dish from the Incubator}\label{c.1.3-remove-the-petri-dish-from-the-incubator}

\begin{figure*}[t]
\centering
\includegraphics[width=0.95\textwidth]{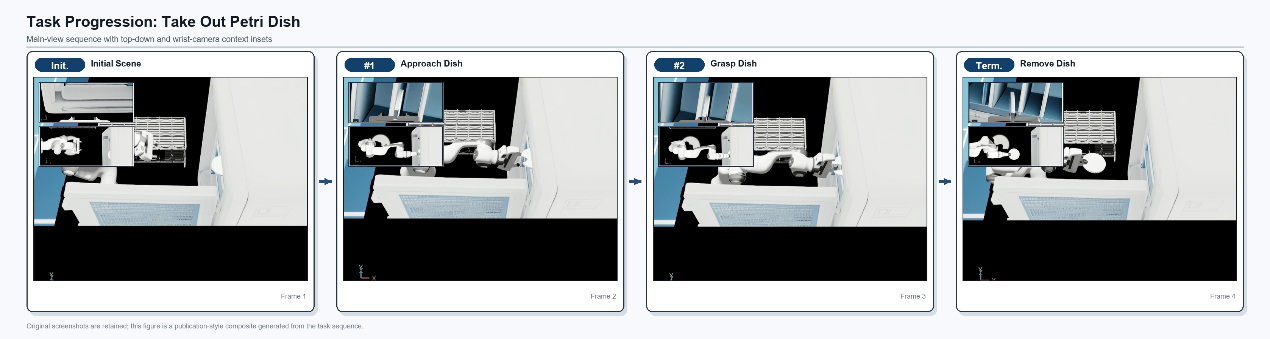}
\caption{Task sequence for Remove the Petri Dish from the Incubator.}
\label{fig:task-29}
\end{figure*}

\par\noindent
\textbf{Description.} This task requires the robotic arm to retrieve a petri dish from an incubator or petri dish storage area. At the start of the task, the petri dish is located inside a container or device; the robotic arm must identify the dish's position and plan an accessible grasping path. The execution process includes: the end-effector moving to the vicinity of the petri dish, descending and aligning with the edge of the dish or the graspable area, closing the gripper to complete the grasp, and then removing the petri dish from its original storage location and lifting it to a safe height.\par
\noindent\textbf{Success Criteria.} The petri dish must be removed steadily without falling, tilting, or colliding severely with surrounding equipment during the process. The task is considered successful if the petri dish leaves its original location, reaches the designated position, and remains above the required height. If the petri dish remains in its original location, falls, or drops below the safety height, the task is deemed a failure.\par

\subsubsection{Place the Petri Dish in the Incubator}\label{c.1.4-place-the-petri-dish-in-the-incubator}

\begin{figure*}[t]
\centering
\includegraphics[width=0.95\textwidth]{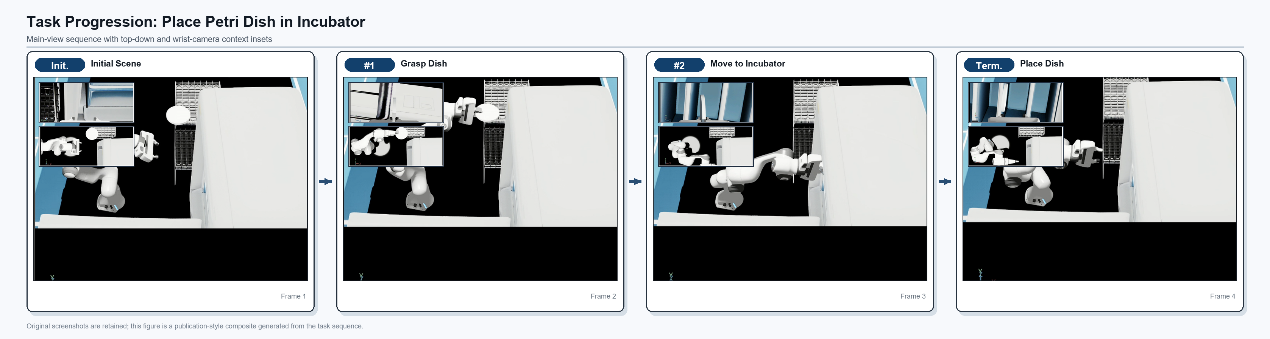}
\caption{Task sequence for Place the Petri Dish in the Incubator.}
\label{fig:task-30}
\end{figure*}

\par\noindent
\textbf{Description.} This task requires the robotic arm to place a petri dish into an incubator or a designated target area. At the start of the task, the petri dish has already been grasped by the robotic arm or is in a manipulable position; the policy must use visual information to locate the target placement area. The execution process includes: the robotic arm moves with the petri dish to a position above the target area, adjusts the end-effector orientation to align the petri dish with the target area, and then gradually lowers the petri dish into the target area and releases it.\par
\noindent\textbf{Success Criteria.} The petri dish must be placed within the designated target area and remain stable after release. For the task to be considered successful, the petri dish's position must meet the target area constraints, its height and orientation must remain within reasonable limits, and it must not fall, slide out, or tilt noticeably within a short period of time. If the petri dish fails to enter the target area or is unstable after release, the task is deemed a failure.\par

\subsection{Equipment Hatch and Opening/Closing Control Tasks}\label{c.2-equipment-hatch-and-openingclosing-control-tasks}

\subsubsection{Close the Centrifuge Lid}\label{c.2.1-close-the-centrifuge-lid}

\begin{figure*}[t]
\centering
\includegraphics[width=0.95\textwidth]{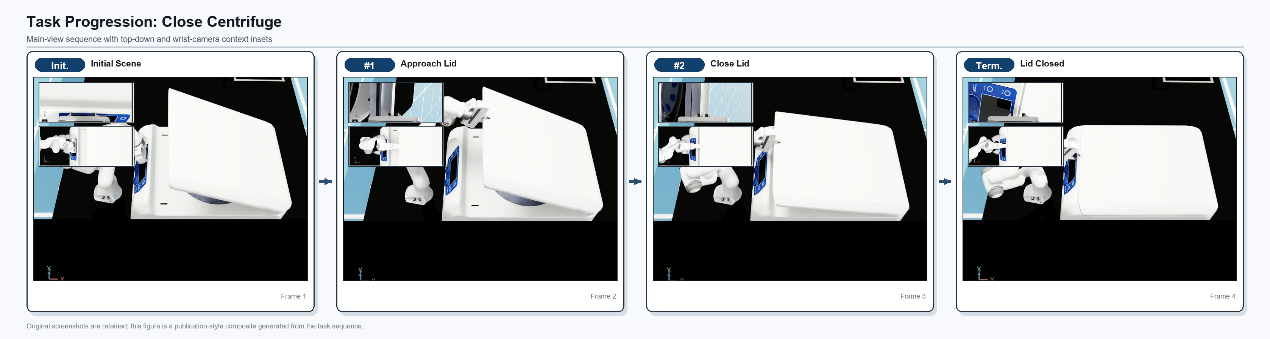}
\caption{Task sequence for Close the Centrifuge Lid.}
\label{fig:task-31}
\end{figure*}

\par\noindent
\textbf{Description.} This task requires the robotic arm to close the lid of an open centrifuge. At the start of the task, the centrifuge lid is open, and the robotic arm must locate the upper surface or a pressable area of the lid. The execution process includes: the end-effector moving above the centrifuge lid, descending to contact the lid's surface, and then applying pressure in the closing direction to cause the lid to move downward around the hinge axis until it approaches the closed position. Finally, the robotic arm pauses briefly to ensure the lid is securely in the closed position.\par
\noindent\textbf{Success Criteria.} The centrifuge lid must be lowered smoothly by the robotic arm; there must be no abnormal collisions or noticeable misalignment during the closing process. For the task to be considered successful, the lid must reach the preset closing threshold and remain stable for a brief period. If the lid is not fully closed, bounces back, or fails to reach the target angle, the task is deemed a failure.\par

\subsubsection{Open the Centrifuge Lid}\label{c.2.2-open-the-centrifuge-lid}

\begin{figure*}[t]
\centering
\includegraphics[width=0.95\textwidth]{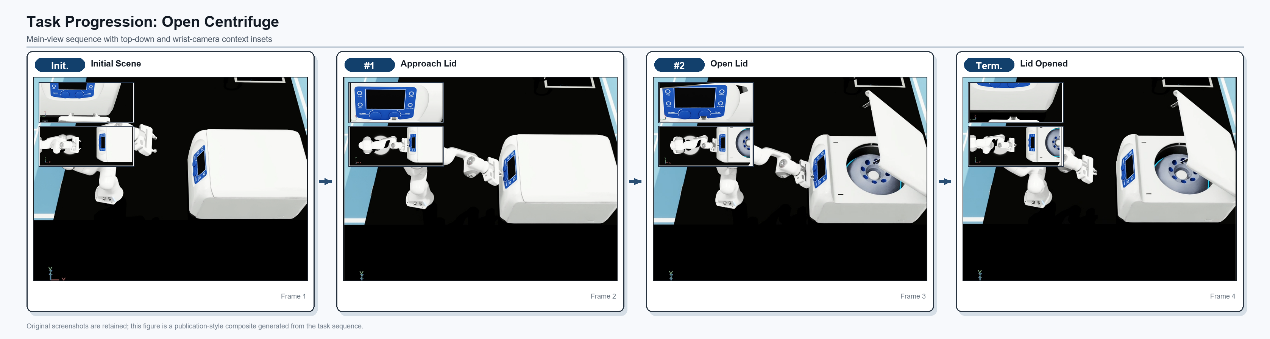}
\caption{Task sequence for Open the Centrifuge Lid.}
\label{fig:task-32}
\end{figure*}

\par\noindent
\textbf{Description.} This task requires the robotic arm to open the centrifuge lid. Rather than pulling the lid open directly, this task triggers the lid to open by pressing the lid-opening button on the centrifuge panel. At the start of the task, the robotic arm must locate the centrifuge panel and the lid-opening button based on visual observations. The execution process includes: the end-effector moving near the button, adjusting its orientation to align with the button, and performing a pressing action toward the button; once the gripper or end-effector enters the virtual button trigger zone, the centrifuge lid mechanism is activated, and the lid opens automatically.\par
\noindent\textbf{Success Criteria.} The robotic arm must accurately contact and activate the lid-opening button without deviating significantly from the button area. The task is considered successful if the centrifuge lid is activated to open and the lid's joint or drive target reaches the preset lid-opening threshold. If the button is not effectively activated or the lid does not reach the open position, the task is deemed a failure.\par

\subsubsection{Open the Water Bath Lid}\label{c.2.3-open-the-water-bath-lid}

\begin{figure*}[t]
\centering
\includegraphics[width=0.95\textwidth]{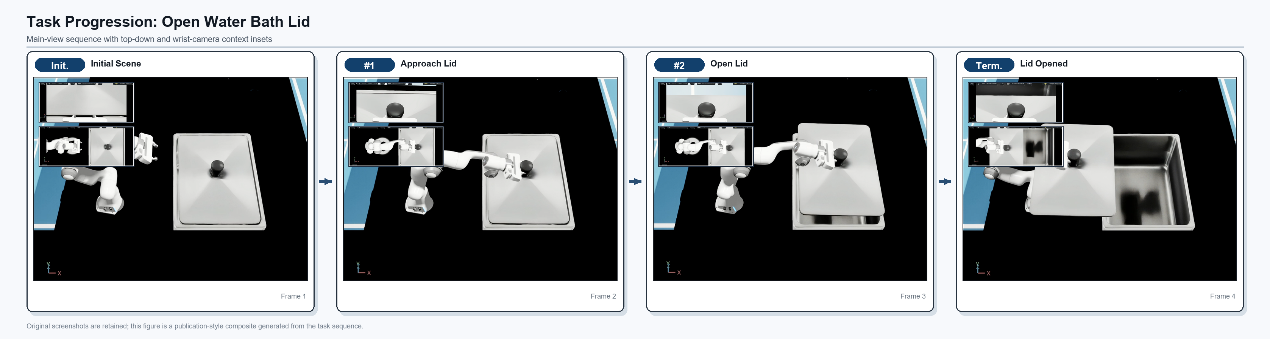}
\caption{Task sequence for Open the Water Bath Lid.}
\label{fig:task-33}
\end{figure*}

\par\noindent
\textbf{Description.} This task requires the robotic arm to open the lid of a water bath. At the start of the task, the lid is closed, and the robotic arm must locate the edge of the lid or the manipulable area. The execution process includes: moving the end-effector near the lid, adjusting its orientation, and making contact with the lid at the point where it can be opened; then applying force in the direction of opening to cause a noticeable displacement of the lid; finally, the robotic arm pauses briefly to ensure that the lid remains open.\par
\noindent\textbf{Success Criteria.} The robotic arm must successfully open the lid of the water bath and displace it sufficiently from its initial position. The task is considered successful if the lid's displacement in the specified direction exceeds the threshold, the change in height meets the criteria for determining that the lid is open, and the lid remains stable for a short period of time. If the lid is not opened, the displacement is insufficient, or the lid closes again after being opened, the task is deemed a failure.\par

\subsubsection{Close the Spectrophotometer Lid}\label{c.2.4-close-the-spectrophotometer-lid}

\begin{figure*}[t]
\centering
\includegraphics[width=0.95\textwidth]{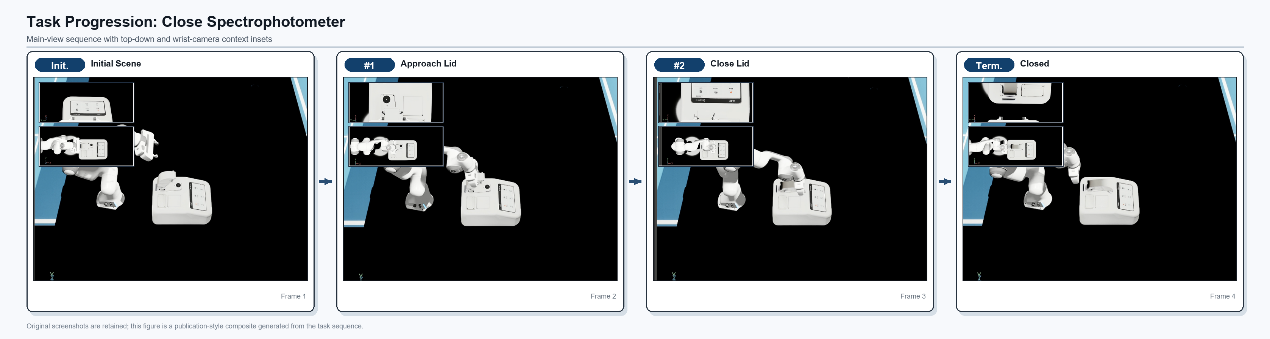}
\caption{Task sequence for Close the Spectrophotometer Lid.}
\label{fig:task-34}
\end{figure*}

\par\noindent
\textbf{Description.} This task requires the robotic arm to close the spectrophotometer's cover. At the start of the task, the spectrophotometer cover is open, and the robotic arm must identify the cover's position and the direction of closure. The execution process includes: the end-effector moving to a position above or to the side of the cover where it can make contact, adjusting its orientation to contact the cover; subsequently pushing the cover in the direction of closure, causing it to rotate around the joint axis to the closed position; finally, the robotic arm withdraws or remains stationary, waiting for the cover to stabilize.\par
\noindent\textbf{Success Criteria.} The spectrophotometer hatch must close smoothly, without any noticeable misalignment or abnormal impact during the closing process. For the mission to be considered successful, the hatch must reach the preset closing threshold angle and remain closed for the specified duration. If the hatch fails to reach the target angle, closes incompletely, or rebounds, the mission is deemed a failure.\par

\subsection{Instrument Placement and Relocation Tasks}\label{c.3-instrument-placement-and-relocation-tasks}

\subsubsection{Place the Centrifuge Tube on the Electronic Balance}\label{c.3.1-place-the-centrifuge-tube-on-the-electronic-balance}

\begin{figure*}[t]
\centering
\includegraphics[width=0.95\textwidth]{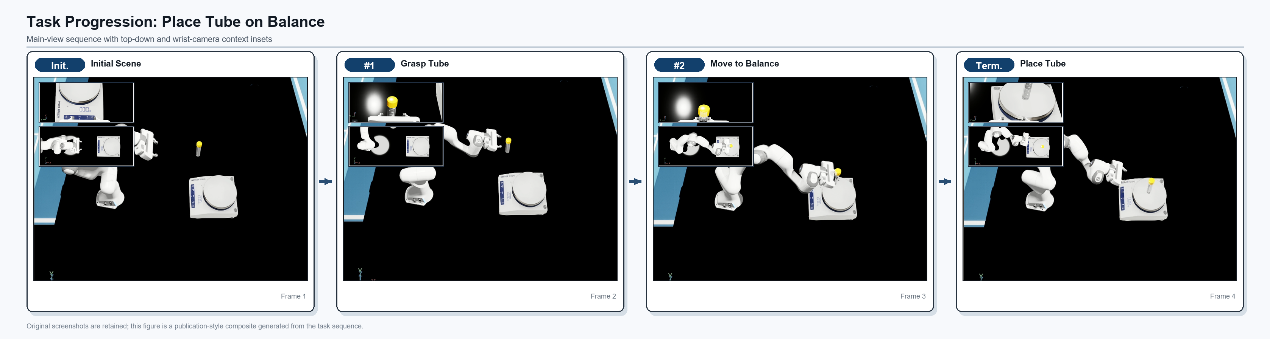}
\caption{Task sequence for Place the Centrifuge Tube on the Electronic Balance.}
\label{fig:task-35}
\end{figure*}

\par\noindent
\textbf{Description.} This task requires the robotic arm to place a centrifuge tube onto the tray of an electronic balance. At the start of the task, the centrifuge tube is located on the lab bench or within the robotic arm's grasping range, and the electronic balance is positioned in the scene as the target instrument. The execution process includes: the robotic arm grasping the centrifuge tube, lifting it, and moving it above the electronic balance; subsequently adjusting the position of the centrifuge tube so that its center aligns with the balance pan; and finally lowering the centrifuge tube and releasing it so that it lands steadily within the pan area.\par
\noindent\textbf{Success Criteria.} The centrifuge tube must be placed within the valid area of the electronic balance tray. The task is considered successful if the centrifuge tube is positioned horizontally within the circular target area of the tray, at a height within the acceptable range, and remains stable for the specified duration. If the tube does not enter the target area, is not within the acceptable height range, or exceeds the time limit, the task is deemed a failure.\par

\subsubsection{Remove the Centrifuge Tube from the Electronic Balance}\label{c.3.2-remove-the-centrifuge-tube-from-the-electronic-balance}

\begin{figure*}[t]
\centering
\includegraphics[width=0.95\textwidth]{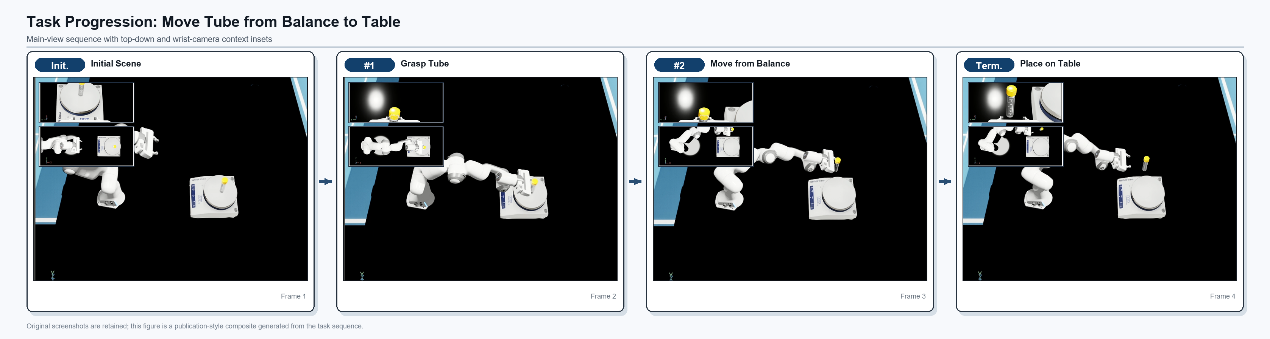}
\caption{Task sequence for Remove the Centrifuge Tube from the Electronic Balance.}
\label{fig:task-36}
\end{figure*}

\par\noindent
\textbf{Description.} This task requires the robotic arm to remove a centrifuge tube from an electronic balance. At the start of the task, the centrifuge tube is located on the electronic balance tray, and the robotic arm must perform a precise grasp within a small target area. The execution process includes: the end-effector moving above the centrifuge tube, descending, and adjusting the position of the gripper to align it with the centrifuge tube; subsequently, closing the gripper to grasp the centrifuge tube, lifting it off the balance tray, and moving it away from the tray area; finally, maintaining a stable posture.\par
\noindent\textbf{Success Criteria.} The centrifuge tube must be successfully removed from the electronic balance by the robotic arm and moved away from the tray area. For the task to be considered successful, the centrifuge tube must be displaced sufficiently from its initial position while maintaining a stable orientation; it must not fall or tilt excessively. If the centrifuge tube remains on the balance, is not effectively moved, or becomes unstable after being grasped, the task is deemed a failure.\par

\subsubsection{Place the Pipette in the Pipette Rack}\label{c.3.3-place-the-pipette-in-the-pipette-rack}

\begin{figure*}[t]
\centering
\includegraphics[width=0.95\textwidth]{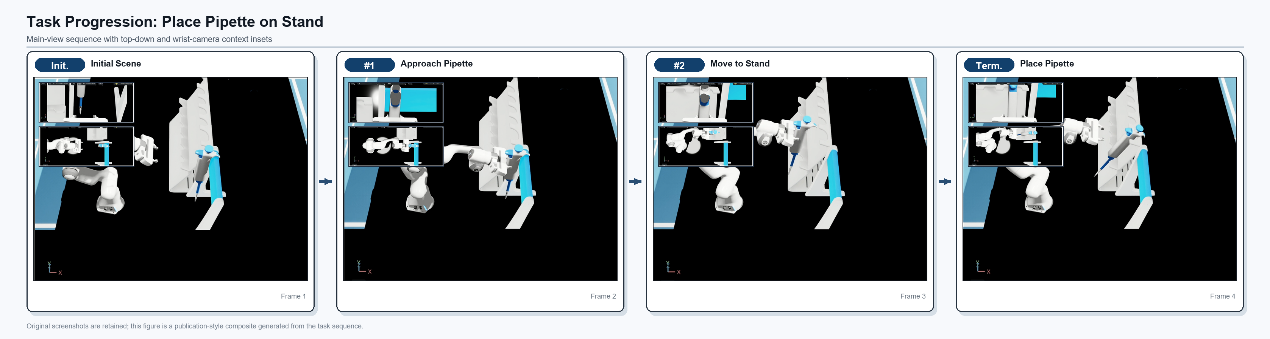}
\caption{Task sequence for Place the Pipette in the Pipette Rack.}
\label{fig:task-37}
\end{figure*}

\begin{figure*}[t]
\centering
\includegraphics[width=0.95\textwidth]{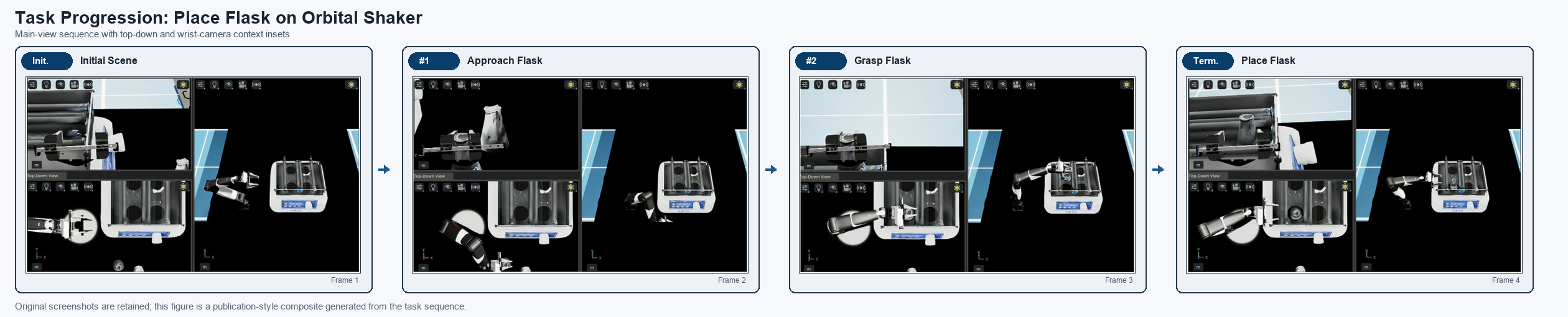}
\caption{Task sequence for Place the Erlenmeyer Flask on the Orbital Shaker.}
\label{fig:task-38}
\end{figure*}

\par\noindent
\textbf{Description.} This task requires the robotic arm to place the pipette onto the pipette rack. At the start of the task, the pipette is located within the robotic arm's grasping range, and the pipette rack, serving as the target support structure, is positioned on the laboratory bench. The execution process includes: the robotic arm grasping the pipette, lifting it, and moving it near the pipette rack; subsequently, adjusting the pipette's orientation based on the holder's spatial position so that the pipette aligns with the placement area on the holder; and finally, lowering the pipette and releasing it to ensure it is securely placed on the holder.\par
\noindent\textbf{Success Criteria.} The pipette must reach the target area of the pipette rack. For the task to be considered successful, the pipette's position must meet the spatial thresholds of the target area, specifically the specified height and horizontal position constraints. If the pipette falls, fails to enter the holder area, or is unstable after release, the task is deemed a failure.\par

\subsubsection{Place the Erlenmeyer Flask on the Orbital Shaker}\label{c.3.4-place-the-erlenmeyer-flask-on-the-shaker}

\par\noindent
\textbf{Description.} This AgileX (Songling) robotic-arm task requires placing an Erlenmeyer flask onto the valid platform region of an orbital shaker. The flask starts within the arm's reachable workspace. The execution proceeds through approaching the flask, establishing a stable grasp, lifting it without collision, moving above the shaker, aligning the flask with the target region, and lowering and releasing it onto the platform.\par
\noindent\textbf{Success Criteria.} The flask must be placed within the valid orbital-shaker platform region and remain stable after release. The task succeeds when the flask center and height meet the target thresholds and no excessive tilt, fall, or displacement occurs during the hold period. Missing the platform, violating the pose thresholds, or losing stability constitutes a failure.\par

\section{Some Additional Features of the Pipette Platform}\label{d-some-additional-features-of-the-pipette-platform}

\subsection{Batch Training Script}\label{d.1-batch-training-script}

In addition to data collection, data transformation, and online evaluation workflows for single tasks, the Pipette Platform also provides batch training scripts for multi-task experiments, enabling the unified launch of training for policy models such as ACT, SmolVLA, and \ensuremath{\pi}0 across 11 wet lab tasks. This design reduces human error caused by repetitive command configuration and allows different models to run under the same set of tasks, data directories, and training entry points, thereby improving the reproducibility of experimental workflows.

The batch training script takes a list of task names as input and automatically concatenates the LeRobot dataset path, model output path, and policy repository identifier based on the task names. The script sets uniform default training parameters for different models; for example, the default batch size for ACT is 32, with 15,000 training steps; SmolVLA uses a default batch size of 8 and trains for 20,000 steps; \(\pi\)0 uses a default batch size of 4 and trains for 20,000 steps. For \(\pi\)0, the script additionally configures parameters such as pre-trained weights, bfloat16, gradient checkpoints, visual encoder freezing, and expert-only training to reduce GPU memory usage and adapt to current hardware conditions.

This script supports both raw data and augmented data training configurations; users can switch between them using the `dataset-version` parameter. Before execution, you can use the `dry-run` mode to print the complete command, which allows you to verify that the data paths, output paths, and model parameters are correct. During training, the script creates separate output directories and log files for each task. If the target directory already exists, you can choose to skip, overwrite, or continue executing the remaining tasks. This mechanism makes long-duration batch training more stable and is suitable for continuously training multiple task models in server or container environments.

\subsection{USD Asset Reference Path Repair}\label{d.2-usd-asset-reference-path-repair}

When building wet lab simulation scenes, USD files often reference external materials, textures, sub-assets, or USDZ packages. If these references use native absolute paths, resources may be lost when the scene is migrated to other machines, servers, or container environments. To improve the portability of asset packages, Pipette Platform provides a tool for checking and repairing USD asset reference paths, which converts absolute resource references in USD files to relative references.

This tool scans for .usd, .usda, and .usdc files in the specified directory and checks the sublayer, reference, payload, and Sdf.AssetPath properties in each USD stage. It determines whether each resource path is absolute or relative and writes the results to a JSON report, helping users identify asset references that still rely on local paths.

For USD files that need to be repaired, the tool will attempt to locate the corresponding USDZ package in the directory containing the current USD file or in a sibling directory, and rewrite the absolute path as a relative package reference. For example, if a USD file references xxx.usdz{[}texture.png{]} on the local path, the tool will search for a matching xxx.usdz file in the current asset directory and rewrite the reference to xxx.usdz{[}texture.png{]}. This way, as long as the USD file and the USDZ package remain in the same asset directory, the scene can properly parse resources on different machines without relying on the absolute paths from the original modeling environment.

\section{Detailed Experimental and Benchmark Settings}\label{app:detailed-experimental-benchmark-settings}

\subsection{Experimental Setup and Training Configuration}\label{app:detailed-experimental-settings}

The experiments validate the Pipette Platform as a benchmark for VLA tasks in wet labs and analyze the impact of simulation-augmented data on different visuomotor policies. All experiments were conducted in the Isaac Sim 5.1 / Isaac Lab 2.3 environment, using a unified sensor configuration, embodiment-specific action space, and task success criteria. Observations include three RGB image streams (top-view, main-view, and wrist-view images) and robot proprioceptive states; actions consist of robotic-arm joint targets and 1-dimensional gripper commands. Franka Panda is used as the default robotic embodiment, while the Erlenmeyer-flask-to-shaker placement task is collected and evaluated with an AgileX robotic arm. Each model is evaluated over 100 episodes per task, with task success rate serving as the performance metric.

We compare ACT, SmolVLA, and \ensuremath{\pi}0 under the same task scenarios, robotic configuration, and success criteria within each task. ACT is used as the behavior-cloning baseline, while SmolVLA and \ensuremath{\pi}0 are VLA policies. We evaluate two data settings: raw demonstrations with 30 human trajectories per task, and raw demonstrations combined with simulation-augmented data. Augmentation is generated through trajectory replay with lighting, camera, speed, and motion perturbations, followed by task-success verification.

\begin{table*}[t]
\centering
\small
\setlength{\tabcolsep}{3pt}
\begin{tabular}{@{}
  >{\raggedright\arraybackslash}p{(\linewidth - 2\tabcolsep) * \real{0.3264}}
  >{\raggedright\arraybackslash}p{(\linewidth - 2\tabcolsep) * \real{0.6669}}@{}}
\toprule\noalign{}
\begin{minipage}[b]{\linewidth}\raggedright
\textbf{Configuration options}
\end{minipage} & \begin{minipage}[b]{\linewidth}\raggedright
\textbf{Settings}
\end{minipage} \\
\midrule\noalign{}
Simulation environment & Isaac Sim 5.1 / Isaac Lab 2.3 \\
Robotic arm & Franka Panda by default; AgileX robotic arm for Erlenmeyer-flask-to-shaker placement \\
Observation space & Top, Main, and Wrist RGB images + 8-dimensional proprioceptive state \\
Action space & Embodiment-specific joint targets + 1-dimensional gripper command \\
Policy query frequency & 10 Hz \\
Control frequency & 30 Hz \\
Initial state & Fixed initial state in the task registry \\
Data settings & Raw demonstration data; Raw demonstration data + simulation-augmented data \\
Number of demonstrations per task & 30 human demonstration trajectories \\
Number of evaluations & 100 episodes per model and per task \\
Evaluation criteria & Task success rate \\
\bottomrule
\end{tabular}
\caption{Experimental setup for policy training and evaluation}
\label{tab:experimental-setup}
\end{table*}

\begin{table*}[t]
\centering
\small
\setlength{\tabcolsep}{3pt}
\begin{tabular}{@{}
  >{\raggedright\arraybackslash}p{(\linewidth - 6\tabcolsep) * \real{0.2704}}
  >{\raggedright\arraybackslash}p{(\linewidth - 6\tabcolsep) * \real{0.2403}}
  >{\raggedright\arraybackslash}p{(\linewidth - 6\tabcolsep) * \real{0.2403}}
  >{\raggedright\arraybackslash}p{(\linewidth - 6\tabcolsep) * \real{0.2403}}@{}}
\toprule\noalign{}
\begin{minipage}[b]{\linewidth}\raggedright
\textbf{Parameters}
\end{minipage} & \begin{minipage}[b]{\linewidth}\raggedright
\textbf{ACT}
\end{minipage} & \begin{minipage}[b]{\linewidth}\raggedright
\textbf{SmolVLA}
\end{minipage} & \begin{minipage}[b]{\linewidth}\raggedright
\(\pi\)\textbf{0}
\end{minipage} \\
\midrule\noalign{}
Batch size & 32 & 8 & 4 \\
Number of training steps & 15000 & 20000 & 20000 \\
Training accuracy & Default & Default & bfloat16 \\
Gradient checkpoints & No & No & Yes \\
Visual encoder freeze & No & No & Yes \\
Pre-trained weights & No & No & lerobot/pi0\_base \\
Expert-only training & No & No & Yes \\
Hardware & RTX 4090 24 GB & RTX 4090 24 GB & RTX 4090 24 GB \\
\bottomrule
\end{tabular}
\caption{Training parameter settings for different policy models}
\label{tab:training-parameters}
\end{table*}

\subsection{Wet-Lab Embodied Task Benchmark}\label{app:detailed-task-benchmark}

The benchmark comprises 12 tasks across sample and culture-consumable handling, equipment-lid control, and instrument placement or relocation. It evaluates visual localization, end-effector pose control, target alignment, contact stability, and long-range execution. Table~\ref{tab:task-benchmark} lists the complete task taxonomy and associated competencies.

\begin{table*}[t]
\centering
\small
\setlength{\tabcolsep}{3pt}
\begin{tabular}{@{}
  >{\raggedright\arraybackslash}p{(\linewidth - 4\tabcolsep) * \real{0.2791}}
  >{\raggedright\arraybackslash}p{(\linewidth - 4\tabcolsep) * \real{0.4268}}
  >{\raggedright\arraybackslash}p{(\linewidth - 4\tabcolsep) * \real{0.2942}}@{}}
\toprule\noalign{}
\begin{minipage}[b]{\linewidth}\raggedright
\textbf{Task Category}
\end{minipage} & \begin{minipage}[b]{\linewidth}\raggedright
\textbf{Specific tasks}
\end{minipage} & \begin{minipage}[b]{\linewidth}\raggedright
\textbf{Key competencies}
\end{minipage} \\
\midrule\noalign{}
\multirow{4}{=}{Sample and Culture Consumables Handling Tasks} & Pick up the test tube & Small object recognition and stable grasping \\
& Pipette-to-petri-dish positioning & Long-distance movement and alignment with the target area \\
& Remove the petri dish from the incubator & Target localization and retrieval in confined spaces \\
& Place the petri dish in the incubator & Alignment and release under spatial constraints \\
\multirow{4}{=}{Equipment Hatch and Opening/Closing Control Tasks} & Close the centrifuge lid & Control of hinge-cover pressing and closing \\
& Open the centrifuge lid & Locating and pressing the small button to open the lid \\
& Open the water bath lid & Lid-structure recognition and lid-opening control \\
& Close the spectrophotometer lid & Closing control for small equipment covers \\
\multirow{4}{=}{Instrument Placement and Relocation Tasks} & Place the centrifuge tube on the electronic balance & Low-height picking and precise placement \\
& Remove the centrifuge tube from the electronic balance & Grasping and obstacle avoidance near workspace boundaries \\
& Place the pipette in the pipette rack & Tool-orientation alignment and stable release \\
& Place the Erlenmeyer flask on the orbital shaker (AgileX/Songling arm) & Vessel transfer and target-platform alignment \\
\bottomrule
\end{tabular}
\caption{Wet-lab manipulation tasks and the key competencies they assess}
\label{tab:task-benchmark}
\end{table*}

\end{document}